\documentclass{article}

\usepackage[main, final]{neurips_2026}

\usepackage[utf8]{inputenc} 
\usepackage[T1]{fontenc}    
\usepackage{hyperref}       
\usepackage{url}            
\usepackage{booktabs}       
\usepackage{amsfonts}       
\usepackage{nicefrac}       
\usepackage{microtype}      
\usepackage{xcolor}         
\usepackage{needspace}

\usepackage{graphicx}
\usepackage{subcaption}
\usepackage{multirow}
\usepackage{enumitem}

\usepackage{amsmath}
\usepackage{amssymb}
\usepackage{mathtools}
\usepackage{amsthm}

\hypersetup{
    colorlinks=true,
    citecolor=blue,  
    linkcolor=red,   
    urlcolor=blue
}

\usepackage{amsmath,amsfonts,bm}

\def\eqref#1{equation~\ref{#1}}

\def\1{\bm{1}}

\DeclareMathAlphabet{\mathsfit}{\encodingdefault}{\sfdefault}{m}{sl}
\SetMathAlphabet{\mathsfit}{bold}{\encodingdefault}{\sfdefault}{bx}{n}

\theoremstyle{plain}

\theoremstyle{definition}

\theoremstyle{remark}

\usepackage{wrapfig}

\title{Improving LLM Final Representations with Inter-Layer Geometry}

\author{%
Tom Ulanovski\thanks{Equal contribution. Listing order is random.} \\
Blavatnik School of Computer Science\\
Tel Aviv University\\
\texttt{tomulanovski@mail.tau.ac.il}
\And
Eyal Blyachman\footnotemark[1] \\
Tel Aviv University\\
\texttt{blyachman1@mail.tau.ac.il}
\And
Maya Bechler-Speicher \\
Meta AI \\
\texttt{mayabs@meta.com}
}

\begin{document}
\maketitle

\begin{abstract}

The standard in LLM-based prediction is to use the final-layer representation as
the input to a downstream predictor. However, intermediate layers may encode
complementary task-relevant signals. Existing approaches therefore either search for the best layer for each task or apply expensive attention-based mechanisms to learn
inter-layer aggregation. In this work, we first show that such complexity is
unnecessary: a lightweight Graph Neural Network over a fully connected graph of
LLM layers is more efficient and achieves significantly stronger predictive performance than existing approaches.
We then introduce the Cayley-Encoder, which further improves both efficiency and predictive performance by replacing the fully connected
graph with a Cayley graph over $SL(2,\mathbb{Z}_n)$. These Cayley graphs provide a mathematically grounded topology that is sparse,
regular by construction, and has low diameter. 
This enables effective communication across layers while constraining the aggregation structure and reducing the risk of GNN overfitting. 
In an evaluation of Cayley-Encoder across 13 tasks and
13 LLMs, Cayley-Encoder consistently
outperforms baselines,  achieving improvements of up to 48 percentage points in accuracy, while introducing at most 0.1\%
additional parameters relative to the LLM size.
We further show that Cayley-Encoder is effective in few-shot regimes. Finally, we show that Cayley-Encoder outperforms LoRA fine-tuning
while operating on the frozen LLM. We conclude with an explainability analysis showing that multiple layers contribute meaningfully to
the final prediction, supporting our hypothesis.

\end{abstract}

\section{Introduction}
\label{sec:introduction}

Large Language Models (LLMs) have become a standard backbone for learning transferable text representations across a broad range of downstream prediction tasks~\citep{vaswani2017attention,devlin2019bert,brown2020language}. A common efficient post-training approach is to use a frozen LLM as a feature extractor and train a lightweight prediction head on top of it. The dominant practice is to use the representation from the final layer of the LLM. This convention implicitly assumes that the last layer provides the most task-relevant representation.

A growing body of analysis shows that language models distribute information across depth. Lower layers tend to encode surface-level and lexical information, middle layers often capture syntactic structure, and upper layers are more strongly associated with semantic abstraction~\citep{tenney2019bert,jawahar2019does,clark-etal-2019-bert}. 
Indeed, recent work has shown that intermediate layers can contain substantial task-relevant signal and may outperform final-layer representations depending on the task~\citep{wallat2021bertnesia,fan2024not,skean2025layer}.
There are approaches that focus on attention-based inter-layer aggregation to form a final representation based on information from all layers~\citep{elnokrashy2024depth}.
However, this design remains anchored to the final layer and introduces additional complexity. Other approaches modify the transformer architecture itself~\citep{pagliardini2024denseformer,xiao2025muddformer}, but require training the architecture from scratch. 

In this work, we show how inter-layer information from a frozen LLM can be aggregated in a way that is both more efficient and more effective than existing approaches, by introducing mathematically grounded geometry over the layers and formulating inter-layer aggregation as a graph learning problem. Our hypothesis is that introducing a structure that does not heavily rely on a single layer as in existing approaches is more beneficial for obtaining the final representation from the LLM. 
This view provides a direct mechanism for modeling interactions among layer representations without modifying the underlying LLM, in a way that is independent of the number of input tokens.
 
First, we show that even a simple fully connected layer graph, where every pair of layers can communicate directly through a lightweight GNN, substantially outperforms both attention mechanisms and selecting the single best-performing layer. 
We then show that both memory efficiency and predictive performance can be further improved by replacing the dense fully connected graph with a special sparse graph. 
While there are many ways to construct sparse graphs, prior work showed that GNNs tend to overfit the underlying graph structure when the structure does not carry information for the predictive task, and that regular graphs are more robust to this overfitting~\citep{bechler2024graph}. Moreover, arbitrary regular graphs may still introduce information bottlenecks and poor communication between nodes~\citep{alon2020bottleneck,wilson2025cayley}. To address these limitations, we introduce the Cayley-Encoder, a geometric approach that replaces the dense fully connected inter-layer graph with a sparse graph with guaranteed advantages. Cayley-Encoder maps layers to a Cayley graph over the Special Linear group $SL(2,\mathbb{Z}_n)$, which is guaranteed to be sparse, regular, and low-diameter, enabling layers to communicate effectively through the GNN. 

We evaluate Fully-Connected Encoder and Cayley-Encoder on 13 diverse downstream tasks, spanning intent detection, sentiment analysis, domain classification, and semantic similarity, across 13 pretrained LLMs from four families. Across this broad evaluation, Cayley-Encoder consistently outperforms baselines, reaching up to 48 percentage points improvement in accuracy, while adding at most 0.1\% parameters relative to the base LLM. We further show that Cayley-Encoder remains effective in few-shot regimes. In addition, Cayley-Encoder outperforms LoRA fine-tuning~\citep{hu2022lora} on most evaluated tasks while keeping the LLM frozen, showing that better use of existing representations can outperform direct adaptation of model weights.

Finally, we analyze the learned aggregation patterns and find that predictions rely on signals from multiple layers roughly evenly, rather than collapsing to a single dominant depth. This supports the benefit of drawing on the full depth of the LLM rather than relying on any single layer. Overall, our findings suggest that stronger representations can be obtained by allowing all layers to communicate effectively in a symmetric way.

Our main contributions are:
\begin{enumerate}[leftmargin=*] 
\item We formulate inter-layer aggregation in frozen LLMs as a graph learning problem and show that a lightweight GNN over a fully connected layer graph already substantially outperforms attention-based aggregation and best-layer optimization. 
\item We introduce Cayley-Encoder, a sparse geometric inter-layer encoder based on Cayley graphs over $SL(2,\mathbb{Z}_n)$, providing a regular, sparse, and low-diameter structure for efficient communication across layers. 
\item We evaluate Cayley-Encoder across 13 diverse downstream tasks and 13 LLMs, showing consistent gains over existing aggregation methods and LoRA fine-tuning while adding at most 0.1\% parameters relative to the underlying LLM. 
\item We show that effective inter-layer communication, rather than the identity or ordering of individual layers, is the primary driver of performance: all layers contribute roughly equally, performance is robust to layer-to-node assignment, and adding layer-position encodings does not help. \end{enumerate}

\section{Related Work}
\label{sec:related_work}

\subsection{Inter-Layer Representations in LLMs}
\begin{figure}[t]
  \centering
  \includegraphics[width=0.8\columnwidth]{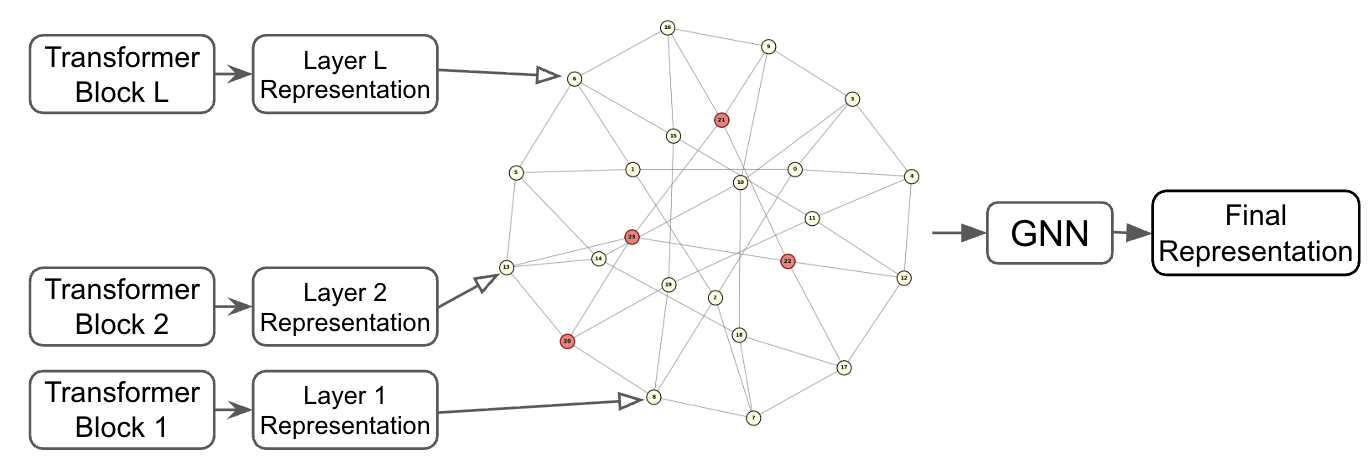} 
  \captionsetup{skip=2pt, belowskip=0pt}
  \vspace{2pt}
  \caption{An overview of Cayley-Encoder.  Layer representations are mapped into nodes in a Cayley Graph over the Special Linear group $SL(2,\mathbb{Z}_n)$, which is then fed into a GNN to learn the final inter-layer representation.}
\label{fig:layer-graph-column}
\end{figure}

Research on transformer depth has revealed that different layers encode different types of information. Lower layers tend to capture surface-level and lexical features, middle layers encode syntactic structure, and upper layers are associated with semantic abstraction~\citep{jawahar2019does,tenney2019bert}. Importantly, intermediate layers can contain task-relevant signals that are absent from the final layer: \citet{wallat2021bertnesia} showed that intermediate layers capture factual knowledge better than the last layer, and \citet{skean2025layer} demonstrated that intermediate representations consistently outperform final-layer ones across downstream tasks. These findings have motivated several approaches for aggregating information across layers. \citet{peters-etal-2018-deep} introduced task-specific scalar weights over BiLSTM layers in ELMo, but this restricts aggregation to a linear mixture that cannot capture non-linear inter-layer interactions. \citet{elnokrashy2024depth} extended this to transformers with Depth-Wise Attention (DWAtt), which attends over layers using a query derived from the final-layer representation. While DWAtt models non-linear interactions, it anchors the aggregation to the last layer, treating earlier layers only as keys and values rather than as equal participants. Other work modifies the transformer architecture itself~\citep{pagliardini2024denseformer,xiao2025muddformer}, but requires training from scratch. Methods such as LayerDrop~\citep{fan2019reducing} operate during training by randomly dropping layers, and are thus complementary to our post-training setting. Our work addresses these limitations by enabling structured inter-layer communication within a frozen LLM, without modifying the architecture or privileging any single layer (Figure~\ref{fig:layer-graph-column}).

\subsection{Graph Neural Networks and Cayley Graphs}

Graph Neural Networks (GNNs) are a family of neural networks designed to learn representations over graph-structured data. GNNs operate by iteratively propagating information between nodes through a process called message passing neural network (MPNN) ~\citep{gilmer2017neural}. Each MPNN layer consists of three operations: (1) each node computes messages to send to its neighbors, (2) each node aggregates the messages it receives, and (3) each node updates its representation by combining the aggregated messages with its previous state.

Many GNN variants have been proposed, differing mostly in how they aggregate information~\citep{kipf2016semi,xu2018powerful}. See \citet{10670406} for a recent survey on GNNs. \citet{bechler2024graph} showed that GNNs tend to overfit the underlying graph structure when the structure does not carry information for the predictive task, and that regular graphs are more robust to this overfitting.
Another limitation of GNNs is over-squashing \citep{alon2020bottleneck}, where information propagating through the graph is compressed into fixed-size node representations.
\citet{deac2022expander} introduced Expander Graph Propagation (EGP) to improve information flow in GNNs. They address over-squashing by leveraging Cayley graphs. 
Cayley graphs over $SL(2, \mathbb{Z}_n)$ with the Margulis generating set are regular and sparse yet highly connected, with logarithmic diameter, allowing any two nodes to communicate in $O(\log |V|)$ steps.
Cayley graphs come in fixed sizes determined by the underlying group. Therefore, EGP truncates nodes to match the input graph size. However, \citet{wilson2025cayley} showed that truncation can reintroduce bottlenecks. They therefore proposed Cayley Graph Propagation (CGP), which instead pads the input graph with virtual nodes to preserve the complete Cayley structure. Virtual nodes only propagate information through the graph and are excluded from the final aggregation.

\citet{mantri2026towards} introduced GLOT, which applies a GNN over a token-similarity graph constructed from the last layer's representations. Unlike our method, GLOT operates within a single layer, requires constructing the graph per input via pairwise token similarity, and produces graphs that grow linearly with input length. Our graph is fixed and determined solely by the number of LLM layers, making it constant-size regardless of input.

\section{Inter-Layer Geometry}\label{sec:methodology}

In this section, we introduce the Fully-Connected and Cayley Encoders.

We consider LLMs with $L$ layers and an input sequence of $T$ tokens. Each layer $\ell \in \{1, \ldots, L\}$ produces a $d$-dimensional hidden representation for each token $\mathbf{H}_\ell \in \mathbb{R}^{T \times d}$. 
We form a \textit{layer representation} $\mathbf{z}_\ell, \ell \in \{1, \ldots, L\}$ by averaging the hidden token representation of each layer:
\begin{equation*}
\mathbf{z}_\ell = \frac{1}{T} \sum_{t=1}^{T} \mathbf{H}_\ell[t, :] \in \mathbb{R}^d
\end{equation*}
Mean-pooling over tokens is a standard approach for obtaining fixed-size sentence representations from LLMs~\citep{reimers2019sentence}. Here, it also ensures that the resulting layer graph has a fixed size determined solely by the number of LLM layers, independent  
  of the input sequence length. We evaluate token-level aggregation as an extension in Section~\ref{sec:token_level}.

We then map each layer representation $\mathbf{z}_\ell$ into nodes of a graph $G = (V, E)$ with nodes $V$ and edges $E$. Each node $v_\ell$ is associated with an initial representation $h_\ell^{(0)} = \mathbf{z}_\ell \in \mathbb{R}^d$. We then apply a GNN over $G$ to produce one final representation $h_G$.
At GNN layer $k$, the representation of node $v$ is updated as follows:
\begin{align*}
    m_v^{(k)} &= \sum_{u \in \mathcal{N}(v)} M^{(k)}\left(h_v^{(k-1)}, h_u^{(k-1)}\right) \\
    h_v^{(k)} &= U^{(k)}\left(h_v^{(k-1)}, m_v^{(k)}\right)
\end{align*}
where $m_v^{(k)}$ is the aggregated message received by node $v$ at layer $k$, $\mathcal{N}(v)$ denotes the set of neighbors of node $v$, $M^{(k)}$ is the message function that computes the contribution from each neighbor, and $U^{(k)}$ is the update function that combines the node's previous representation with its aggregated messages. Both $M^{(k)}$ and $U^{(k)}$ are learnable functions. To obtain a graph-level representation, the final node representations are pooled; the pooling type, mean or sum, is selected as a hyperparameter.

\textbf{Fully-Connected Encoder} In the Fully-Connected Encoder (FC-Encoder) the layer representations are mapped to nodes in a fully connected graph, where every layer is connected to every other layer. This structure is dense, with $\binom{L}{2}$ undirected edges. Note that this corresponds to attention between layers only when the GNN computes attention weights over edges, as in GAT~\citep{velickovic2018graph}; the GNN architectures we evaluate, GIN \citep{xu2018powerful} and GCN \citep{kipf2016semi} (Section~\ref{sec:experiments}), aggregate messages without such weights.

\textbf{Cayley-Encoder} In the Cayley-Encoder, layer representations are mapped to nodes of a Cayley Graph over $SL(2, \mathbb{Z}_n)$.
A Cayley graph is a graph that describes the abstract structure of a mathematical group.
Here we focus on the special linear group $SL(2, \mathbb{Z}_n)$, with elements that are defined as $2 \times 2$ matrices with integer entries modulo $n$ and unit determinant.

The resulting graph is 4-regular and has logarithmic diameter, enabling efficient global communication during message passing (see Appendix~\ref{app:cayley} for construction details).
The number of nodes in the graph $|V_n|$ is determined by the formula:

\begin{equation*}
|V_n| = n^3 \prod_{\text{prime } p | n} \left(1 - \frac{1}{p^2}\right)
\end{equation*}

To maintain the graph's symmetry and expansion properties, we choose the smallest $n$ such that $|V_n| \geq L$. We map the $L$ layer representations randomly to $L$ nodes of the Cayley graph. When $|V_n| > L$, the remaining $|V_n| - L$ nodes are initialized as virtual nodes with zero vectors as their representations.
Following \citet{wilson2025cayley}, we use only the representations from the non-virtual nodes for the final hidden representation used for the prediction task.

While the graph is fixed for a given LLM, the encoder is trained separately for each task, so it does not assume that the roles of the layers are the same across tasks. The same connectivity can be used differently for different tasks. In the next section, we demonstrate the effectiveness of Cayley-Encoder and FC-Encoder through extensive evaluation. 

\section{Experiments}
\label{sec:experiments}

In this section we evaluate FC-Encoder and Cayley-Encoder\footnote{Code: \url{https://github.com/eyali123/ILGE}}. 
We evaluate on 13 diverse tasks from the MTEB benchmark~\citep{muennighoff2023mteb}, spanning intent detection, sentiment analysis, domain classification, and semantic textual similarity. Beyond standard full-data evaluation, we assess zero-shot transfer, few-shot learning with 1--1024 samples per label, scaling behavior across LLM sizes from 14M to 14B parameters in two LLM families, comparison with LoRA fine-tuning, token-level inter-layer aggregation, and an analysis of layer importance.

\paragraph{Datasets.} We use 13 tasks from the MTEB benchmark~\citep{muennighoff2023mteb}. Five are classification tasks: Banking77 (77 intent classes)~\citep{casanueva-etal-2020-efficient}, Emotion (6 emotion categories)~\citep{saravia-etal-2018-carer}, MTOPDomain and MTOPIntent (task-oriented dialogue)~\citep{li-etal-2021-mtop}, and Poem Sentiment (4 sentiment classes)~\citep{sheng2020investigating}. For each, we train on the training split and evaluate on the test split. The remaining 8 are semantic textual similarity (STS) tasks. We train on STSBenchmark~\citep{huggingface:dataset:stsb_multi_mt} and evaluate on its test split, then perform zero-shot transfer to STS12--16, BIOSSES, and SICK-R~\citep{10.5555/2387636.2387697, Agirre2013SEM2S, agirre2014semeval, agirre2015semeval, agirre-etal-2016-semeval,10.1093/bioinformatics/btx238, marelli-etal-2014-sick}.

\paragraph{Setup.}

We compare both encoders against a diverse set of baselines: \textbf{Last-Layer}, the last-layer representation as in \citet{skean2025layer}; \textbf{Best-Layer}, the best-performing layer selected by evaluating each layer independently; \textbf{Weighted}, a learned weighted sum of layer representations, similar to \citet{peters-etal-2018-deep}; \textbf{MLP Last-Layer}, a trained MLP over the last-layer representations; \textbf{DWAtt}, a projection to 256 dimensions followed by DWAtt \citep{elnokrashy2024depth}, where the projection reduces the large parameter count of applying DWAtt directly to layer representations (results without the projection are in Appendix~\ref{app:dwatt}); and \textbf{Deepset}, a Deepset over layer representations \citep{zaheer2017deep}.

 We evaluate on Pythia-410M (25 layers, 1024-dim) \citep{biderman2023pythia}, Gemma2-2B (27 layers, 2304-dim) \citep{gemma_2024}, Llama3-8B (33 layers, 4096-dim) \citep{llama3modelcard}, and Qwen3-8B (37 layers, 4096-dim) \citep{yang2025qwen3}. In addition, Section~\ref{sec:scaling} evaluates six additional Pythia sizes and three additional Qwen3 sizes, for a total of 13 LLMs. All base models remain frozen during training.
 For the first set of experiments, we use the layer representations obtained by mean-pooling over all token representations for each layer. These pooled representations serve as the input to all methods. In this setting the input size is constant and derived from the number of layers in the LLM. We further evaluate FC-Encoder and Cayley-Encoder on all token representations across all layers, making the input size dependent on the input sequence length to the LLM. 

For classification tasks, we train the encoders jointly with a linear classifier head using cross-entropy loss. For STS tasks, we encode each sentence in a pair through the frozen LLM and our encoders, compute the cosine similarity between the resulting representations, and minimize Mean Squared Error (MSE) loss with respect to the ground-truth similarity score.
We use the Adam optimizer~\citep{kingma2014adam} with learning rate and weight decay selected via Optuna~\citep{akiba2019optuna} hyperparameter optimization on the validation set.
For FC-Encoder and Cayley-Encoder, we evaluate GIN \citep{xu2018powerful} and GCN \citep{kipf2016semi} architectures as hyperparameters. Trainable parameter counts and full hyperparameter details are in Appendix~\ref{app:param-counts} and~\ref{app:impl}. 
Following the standard MTEB evaluation protocol~\citep{muennighoff2023mteb}, we report single-run results for each configuration.

\subsection{Results}

Tables~\ref{tab:results} and~\ref{tab:results_sts} present the results. Cayley-Encoder achieves the best performance in 14 out of 20 classification configurations, 3 out of 4 supervised STS configurations, and 17 out of 28 zero-shot STS transfer configurations, making it the strongest method overall.
For classification, Cayley-Encoder achieves average gains of 33 percentage points over Last-Layer and 25 percentage points over Best-Layer, which itself requires evaluating every layer independently. The largest gains appear on the Emotion and Poem Sentiment tasks, with improvements of up to 48 percentage points over Last-Layer. For STS, Cayley-Encoder achieves the highest score in 20 out of 32 configurations, with the largest margins on STS13 and STS14. Comparing to multi-layer aggregation approaches, Cayley-Encoder outperforms DWAtt on all 13 tasks while using 3--5$\times$ fewer parameters (Table~\ref{tab:t-param-counts}). MLP Last-Layer achieves gains on classification but provides no consistent benefit for STS. Both encoders outperform Deepset in most configurations, indicating that inter-layer connectivity improves aggregation beyond permutation-invariant pooling. FC-Encoder, while on par with Cayley-Encoder in some configurations and particularly on smaller LLMs for classification, outperforms DWAtt in most configurations, demonstrating that structured aggregation can be sufficient without attention mechanisms.
 \begin{table*}[!htbp]
  \centering
  \caption{Performance comparison across 4 LLMs on 5 classification tasks. Cayley-Encoder achieves the best score in 14 out of 20 configurations, with its advantage most pronounced on larger LLMs. In all but three cases, FC-Encoder and Cayley-Encoder achieve the top
  two scores. We highlight in bold the best result and in \textcolor{blue}{blue} the second best per column.}
  \resizebox{\textwidth}{!}{
  \begin{tabular}{llccccc}
  \toprule
  Base Model & Method & Banking77 & Emotion & MTOPDomain & MTOPIntent & PoemSentiment \\
  \midrule
  \multirow{8}{*}{Pythia-410M}
  & Last-Layer & 61.17 & 33.48 & 80.88 & 66.97 & 42.40 \\
  & Best-Layer & 66.67 & 35.02 & 83.78 & 71.18 & 45.67 \\
  & MLP Last-Layer & 83.84 & 33.99 & 96.97 & 83.75 & \textcolor{blue}{75.00} \\
  & Weighted & 58.50 & 28.26 & 79.79 & 61.68 & 42.60 \\
  & DWAtt & 83.23 & 58.60 & 98.03 & 91.66 & 70.87 \\
  & Deepset & 84.23 & 47.89 & 97.59 & 92.21 & 73.37 \\
  & FC-Encoder & \textbf{90.65} & \textbf{75.61} & \textcolor{blue}{98.68} & \textbf{95.04} & \textbf{75.77} \\
  & Cayley-Encoder & \textcolor{blue}{89.43} & \textcolor{blue}{73.83} & \textbf{98.77} & \textcolor{blue}{94.72} & 70.87 \\
  \midrule
  \multirow{8}{*}{Gemma2-2B}
  & Last-Layer & 62.16 & 26.89 & 80.79 & 68.02 & 35.67 \\
  & Best-Layer & 72.31 & 31.27 & 87.09 & 75.36 & 42.79 \\
  & MLP Last-Layer & 87.47 & 59.05 & 98.37 & 92.73 & 71.63 \\
  & Weighted & 65.40 & 29.94 & 85.17 & 73.20 & 39.62 \\
  & DWAtt & 88.82 & 61.22 & 98.39 & 90.93 & 67.79 \\
  & Deepset & 90.03 & \textcolor{blue}{78.77} & 98.92 & 94.37 & \textcolor{blue}{78.56} \\
  & FC-Encoder & \textcolor{blue}{92.39} & \textbf{79.90} & \textcolor{blue}{99.07} & \textcolor{blue}{95.34} & 77.12 \\
  & Cayley-Encoder & \textbf{92.58} & 69.60 & \textbf{99.16} & \textbf{96.43} & \textbf{83.27} \\
  \midrule
  \multirow{8}{*}{Llama3-8B}
  & Last-Layer & 68.25 & 34.23 & 84.42 & 73.39 & 40.96 \\
  & Best-Layer & 71.93 & 38.42 & 89.01 & 78.17 & 47.02 \\
  & MLP Last-Layer & 86.70 & 67.67 & 98.58 & 92.09 & 75.00 \\
  & Weighted & 66.63 & 27.94 & 83.85 & 71.78 & 37.79 \\
  & DWAtt & 90.04 & 66.55 & 97.97 & 92.41 & 75.00 \\
  & Deepset & 87.62 & 71.04 & 98.77 & 95.43 & 77.02 \\
  & FC-Encoder & \textcolor{blue}{92.38} & \textcolor{blue}{71.64} & \textcolor{blue}{98.99} & \textcolor{blue}{96.19} & \textcolor{blue}{77.98} \\
  & Cayley-Encoder & \textbf{92.85} & \textbf{73.43} & \textbf{99.03} & \textbf{96.46} & \textbf{79.04} \\
  \midrule
  \multirow{8}{*}{Qwen3-8B}
  & Last-Layer & 48.12 & 30.18 & 71.79 & 56.67 & 37.31 \\
  & Best-Layer & 70.98 & 36.36 & 87.88 & 76.94 & 45.67 \\
  & MLP Last-Layer & 81.37 & 56.95 & 98.09 & 87.10 & 68.65 \\
  & Weighted & 60.51 & 28.52 & 81.71 & 53.63 & 43.08 \\
  & DWAtt & 91.22 & 68.31 & 98.97 & 92.10 & 73.94 \\
  & Deepset & 85.71 & 69.25 & 97.92 & 53.11 & 71.15 \\
  & FC-Encoder & \textcolor{blue}{91.90} & \textcolor{blue}{69.66} & \textbf{99.02} & \textcolor{blue}{95.60} & \textcolor{blue}{80.48} \\
  & Cayley-Encoder & \textbf{92.34} & \textbf{75.70} & \textcolor{blue}{98.99} & \textbf{95.99} & \textbf{85.38} \\
  \bottomrule
  \end{tabular}}
  \label{tab:results}
  \end{table*}

\begin{table*}[htbp] \centering \caption{Performance comparison across 4 LLMs on 8 STS tasks (Spearman correlation $\times 100$). Cayley-Encoder achieves the highest score in 20 out of 32 configurations. We highlight in bold the best result and in \textcolor{blue}{blue} the second best per column.} \resizebox{\textwidth}{!}{ \begin{tabular}{llcccccccc} \toprule Base Model & Method & STSBenchmark & STS12 & STS13 & STS14 & STS15 & STS16 & BIOSSES & SICK-R \\ \midrule \multirow{8}{*}{Pythia-410M} & Last-Layer & 39.12 & 46.96 & 47.00 & 41.45 & 49.32 & 50.37 & 67.30 & 52.55 \\ & Best-Layer & 53.53 & 50.62 & \textcolor{blue}{59.27} & 51.61 & \textcolor{blue}{65.59} & \textcolor{blue}{58.02} & \textbf{74.80} & \textbf{58.26} \\ & MLP Last-Layer & 25.54 & 42.14 & 36.16 & 33.28 & 37.75 & 39.84 & 59.68 & 43.23 \\ & Weighted & 41.81 & 47.12 & 49.49 & 44.21 & 53.72 & 52.76 & \textcolor{blue}{67.79} & 55.09 \\ & DWAtt & 49.40 & \textcolor{blue}{52.51} & 44.57 & 47.46 & 52.51 & 44.80 & 45.29 & 52.43 \\ & Deepset & 39.48 & 52.11 & 42.68 & 43.99 & 48.32 & 40.48 & 44.72 & 44.84 \\ & FC-Encoder & \textcolor{blue}{54.20} & 50.13 & 53.00 & \textcolor{blue}{51.76} & 63.94 & 57.11 & 58.65 & \textcolor{blue}{57.14} \\ & Cayley-Encoder & \textbf{55.84} & \textbf{55.97} & \textbf{60.25} & \textbf{56.65} & \textbf{66.99} & \textbf{58.63} & 56.59 & 55.34 \\ \midrule \multirow{8}{*}{Gemma2-2B} & Last-Layer & 36.82 & 33.70 & 45.60 & 37.58 & 50.22 & 51.40 & 58.44 & 43.01 \\ & Best-Layer & 52.97 & 43.02 & \textcolor{blue}{58.56} & 52.43 & 65.49 & 58.89 & \textbf{72.02} & 57.24 \\ & MLP Last-Layer & 40.72 & 27.36 & 36.93 & 31.32 & 52.38 & 51.16 & 54.10 & 40.19 \\ & Weighted & 40.52 & 35.66 & 45.58 & 38.14 & 54.88 & 54.91 & \textcolor{blue}{65.27} & 46.67 \\ & DWAtt & 53.37 & 51.86 & 48.14 & 55.29 & 67.73 & 51.11 & 57.08 & 55.36 \\ & Deepset & 36.67 & 43.97 & 38.97 & 35.05 & 50.51 & 47.34 & 31.64 & 42.22 \\ & FC-Encoder & \textbf{62.86} & \textbf{61.83} & 55.43 & \textcolor{blue}{63.89} & \textbf{74.68} & \textbf{62.96} & 66.02 & \textcolor{blue}{60.12} \\ & Cayley-Encoder & \textcolor{blue}{61.62} & \textcolor{blue}{60.38} & \textbf{63.98} & \textbf{64.59} & \textcolor{blue}{73.94} & \textcolor{blue}{62.69} & 64.07 & \textbf{60.32} \\ \midrule \multirow{8}{*}{Llama3-8B} & Last-Layer & 45.63 & 32.65 & 52.66 & 42.88 & 58.07 & 54.32 & \textcolor{blue}{67.00} & 51.16 \\ & Best-Layer & 56.51 & 47.21 & \textcolor{blue}{60.09} & 54.62 & 66.89 & \textcolor{blue}{59.04} & \textbf{72.83} & 56.96 \\ & MLP Last-Layer & 48.91 & 34.88 & 54.80 & 44.08 & 61.19 & 52.55 & 55.35 & 46.62 \\ & Weighted & 45.77 & 32.86 & 59.46 & 54.90 & 67.30 & 58.97 & 52.82 & 56.91 \\ & DWAtt & 52.85 & 51.47 & 56.26 & 58.63 & 66.00 & 48.86 & 46.88 & 51.45 \\ & Deepset & 42.31 & 39.94 & 52.80 & 52.00 & 56.12 & 39.50 & 32.96 & 46.60 \\ & FC-Encoder & \textcolor{blue}{59.33} & \textcolor{blue}{55.19} & 53.51 & \textcolor{blue}{59.88} & \textcolor{blue}{73.69} & 57.25 & 62.81 & \textcolor{blue}{57.51} \\ & Cayley-Encoder & \textbf{63.05} & \textbf{65.31} & \textbf{63.53} & \textbf{70.17} & \textbf{76.96} & \textbf{63.13} & 55.32 & \textbf{60.74} \\ \midrule \multirow{8}{*}{Qwen3-8B} & Last-Layer & 29.26 & 58.45 & 37.48 & 33.88 & 47.65 & 46.36 & 51.83 & 41.07 \\ & Best-Layer & 54.20 & 58.45 & \textcolor{blue}{59.74} & 51.52 & 62.63 & \textbf{60.03} & \textbf{69.67} & \textbf{58.11} \\ & MLP Last-Layer & 26.36 & 45.20 & 35.55 & 31.01 & 47.06 & 36.56 & 53.64 & 40.93 \\ & Weighted & 38.45 & 52.03 & 45.05 & 38.63 & 57.53 & 49.71 & 59.45 & 50.34 \\ & DWAtt & 38.63 & 44.96 & 41.00 & 40.56 & 53.98 & 45.68 & 54.26 & 42.05 \\ & Deepset & 39.53 & 49.10 & 33.01 & 32.24 & 46.16 & 48.54 & 57.05 & 44.74 \\ & FC-Encoder & \textcolor{blue}{54.24} & \textbf{64.87} & 49.30 & \textcolor{blue}{59.51} & \textcolor{blue}{69.34} & 57.47 & \textcolor{blue}{65.11} & 56.20 \\ & Cayley-Encoder & \textbf{58.69} & \textcolor{blue}{63.52} & \textbf{64.81} & \textbf{68.85} & \textbf{73.69} & \textcolor{blue}{58.28} & 54.77 & \textcolor{blue}{57.89} \\ \bottomrule \end{tabular}} \label{tab:results_sts} \end{table*}

\subsection{Effectiveness Across LLM Sizes}
\label{sec:scaling}
\begin{figure*}[!htbp]
\centering
    \begin{subfigure}[b]{0.19\textwidth}
        \centering
        \includegraphics[width=\textwidth]{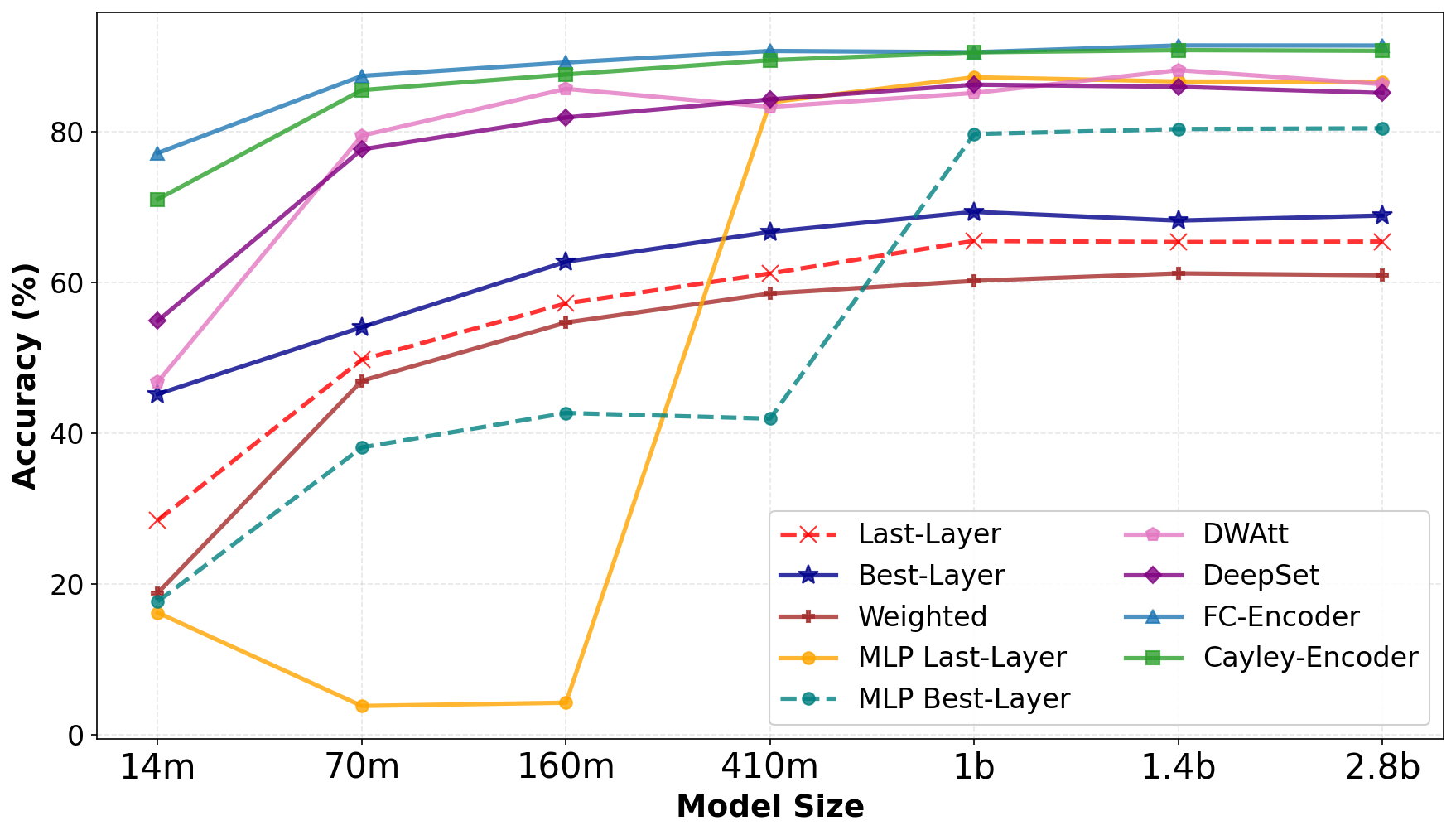}
        \caption{Banking77}
    \end{subfigure}
    \hfill
    \begin{subfigure}[b]{0.19\textwidth}
        \centering
        \includegraphics[width=\textwidth]{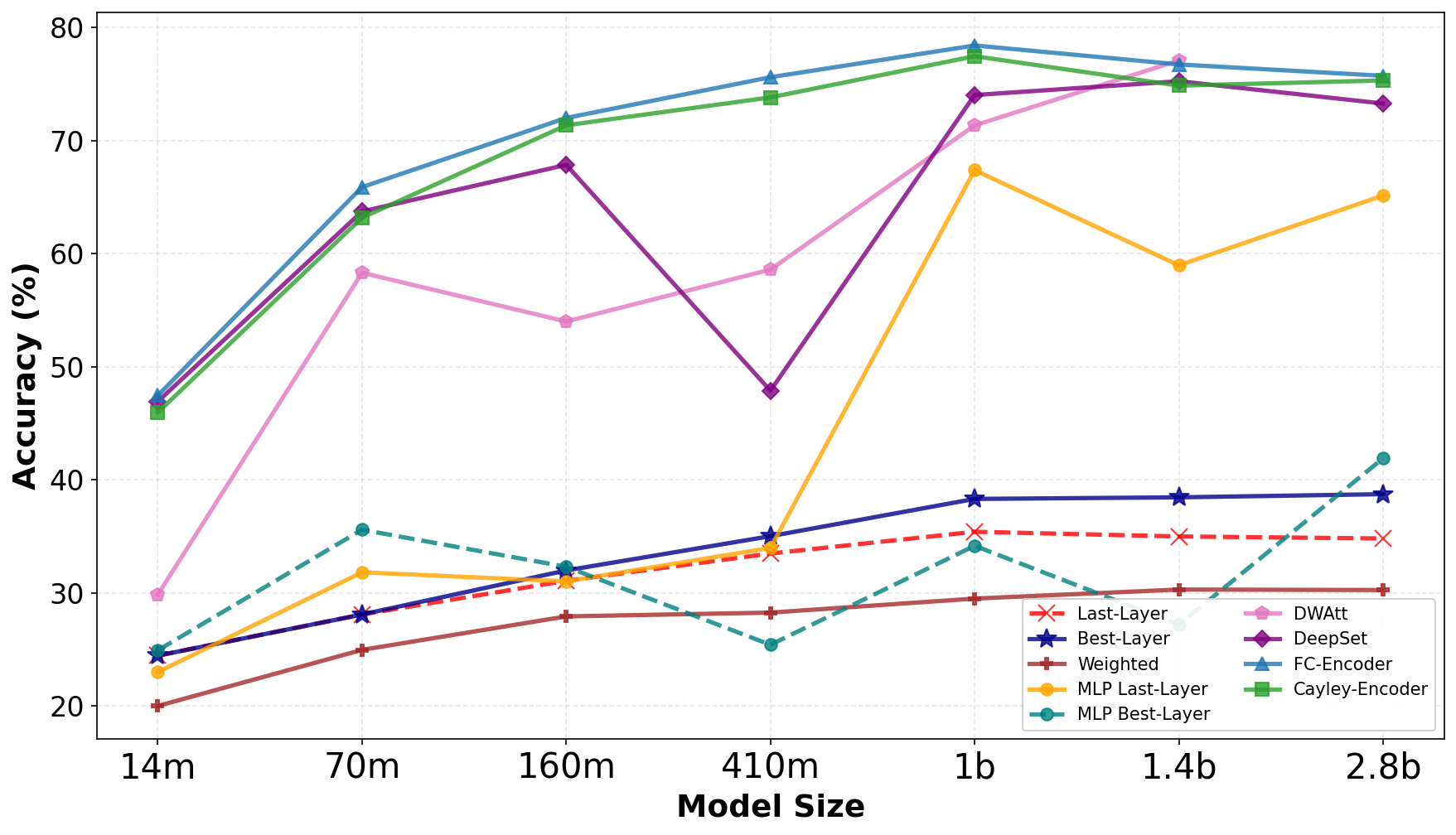}
        \caption{Emotion}
    \end{subfigure}
    \hfill
    \begin{subfigure}[b]{0.19\textwidth}
        \centering
        \includegraphics[width=\textwidth]{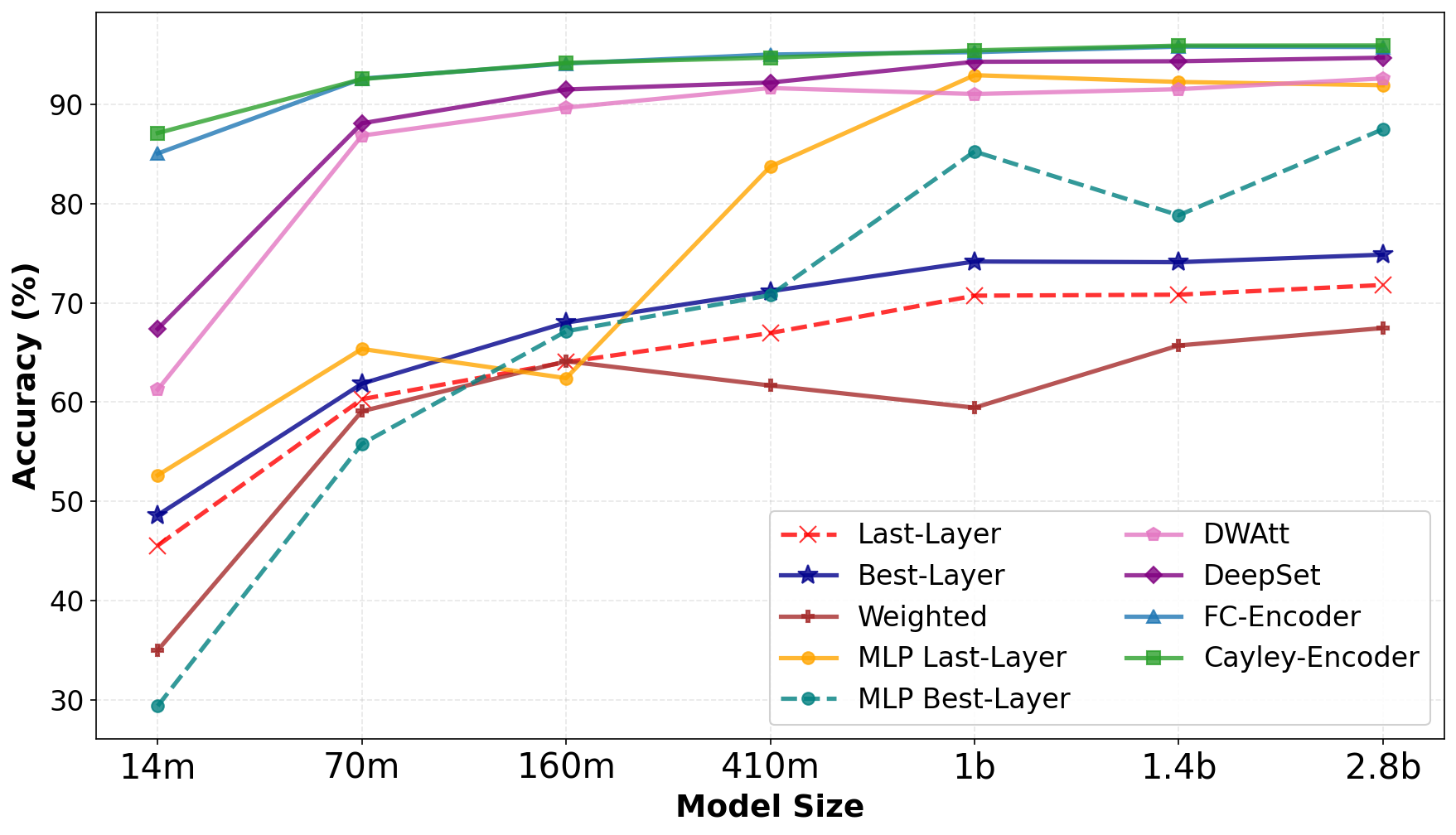}
        \caption{MTOPIntent}
    \end{subfigure}
    \hfill
    \begin{subfigure}[b]{0.19\textwidth}
        \centering
        \includegraphics[width=\textwidth]{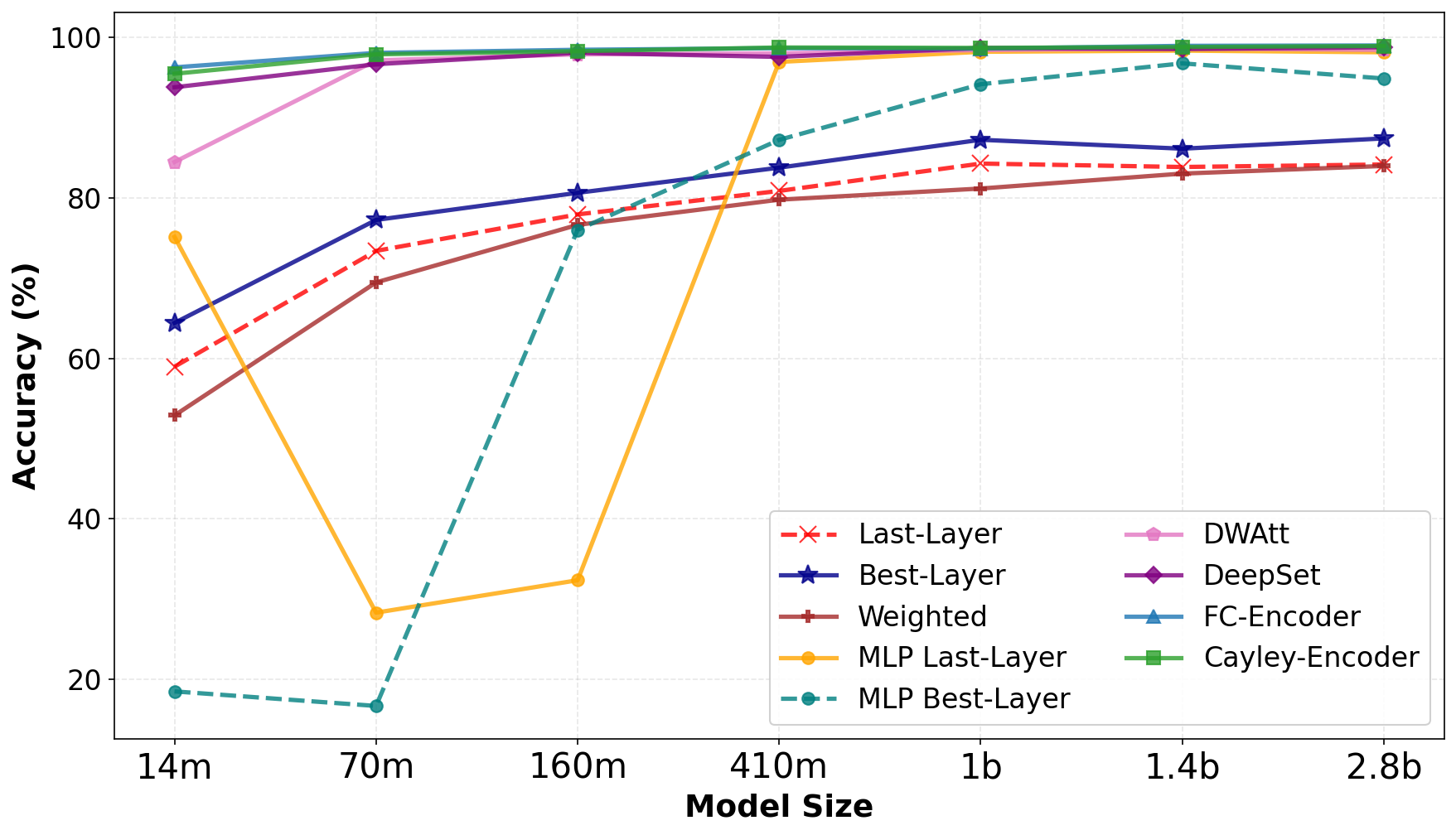}
        \caption{MTOPDomain}
    \end{subfigure}
    \hfill
    \begin{subfigure}[b]{0.19\textwidth}
        \centering
        \includegraphics[width=\textwidth]{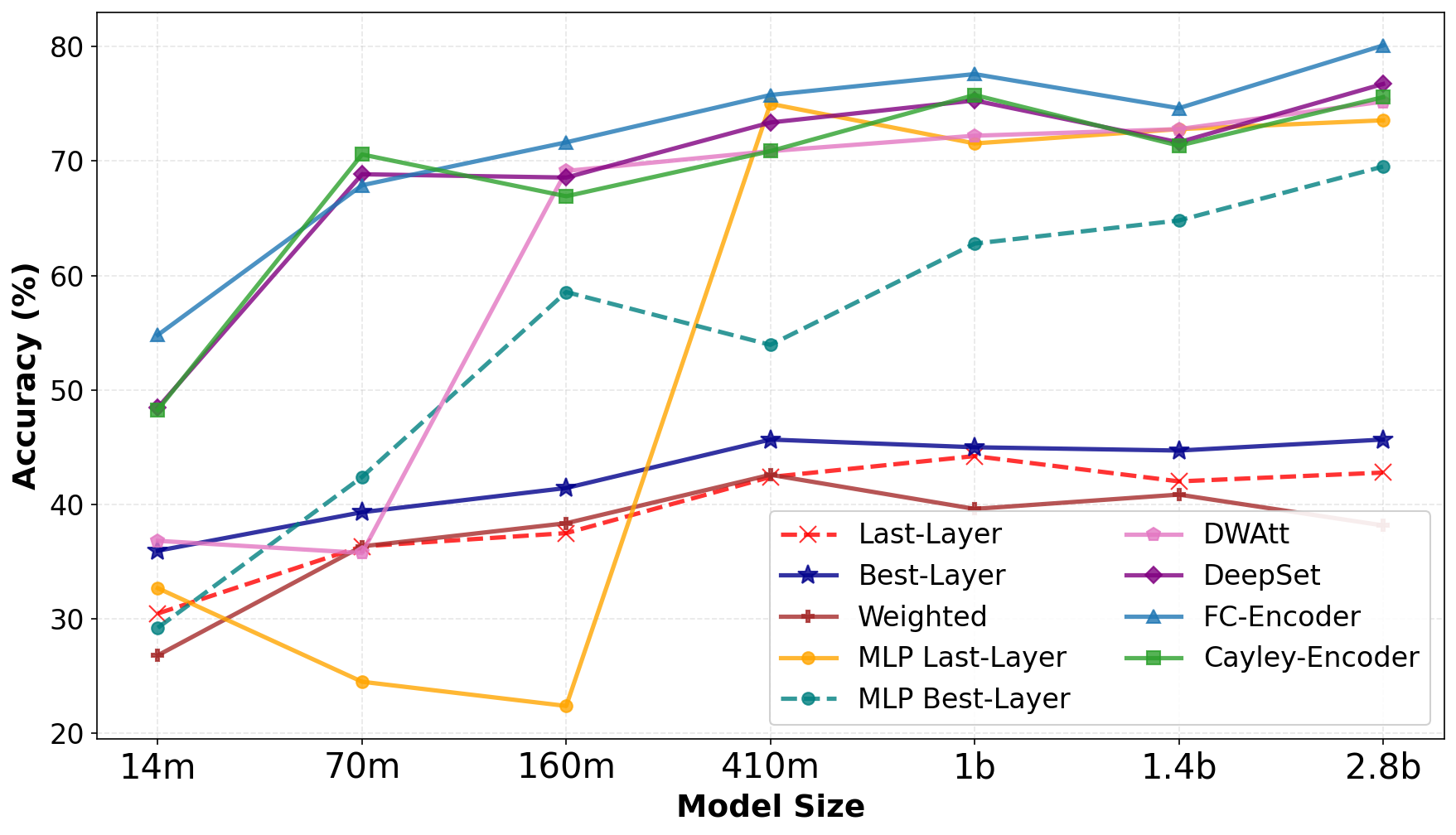}
        \caption{PoemSentiment}
    \end{subfigure}
    \caption{Performance across the Pythia models ranging from 14M to 2.8B parameters. Both encoders outperform Last-Layer and Best-Layer across all model sizes, with smaller LLMs achieving performance comparable to much larger ones.}
    \label{fig:model_scaling}
      \vspace{-1.0em}
\end{figure*}

We evaluate whether the benefits of our encoders persist across LLM sizes. We conduct this evaluation on the classification tasks using the Pythia suite~\citep{biderman2023pythia}: 14M, 70M, 160M, 410M, 1B, 1.4B, and 2.8B parameters. Using a single model family neutralizes confounding effects due to architectural differences between LLM families. Figure~\ref{fig:model_scaling} shows performance across Pythia sizes. Both FC-Encoder and Cayley-Encoder consistently outperform all baselines across all Pythia sizes, with gains of 15--40 percentage points over Last-Layer and Best-Layer maintained throughout. Notably, smaller LLMs equipped with our encoders can match the performance of much larger ones: Pythia-14M already achieves 95--96\% accuracy on MTOPDomain, just 2--4 percentage points below Pythia-2.8B. This suggests that our approach can be particularly valuable in resource-constrained settings, extracting strong performance from small models by better utilizing their existing layer representations.
To verify that these trends are not specific to one family, we repeat the analysis on the Qwen3 family~\citep{yang2025qwen3} with 1.7B, 4B, 8B, and 14B parameters (Appendix~\ref{app:qwen_sizes}). Both encoders outperform Last-Layer by 35--41 percentage points on average at all sizes. Cayley-Encoder outperforms FC-Encoder on average at every size, and its advantage is larger for the two largest LLMs (8B: +2.35, 14B: +1.47) than for the smaller ones (1.7B: +0.74, 4B: +0.31).

\subsection{Few-Shot Learning}
\label{sec:fewshot}

\begin{figure*}[!htbp]
  \centering
      \begin{subfigure}[b]{0.19\textwidth}
          \centering
          \includegraphics[width=\textwidth]{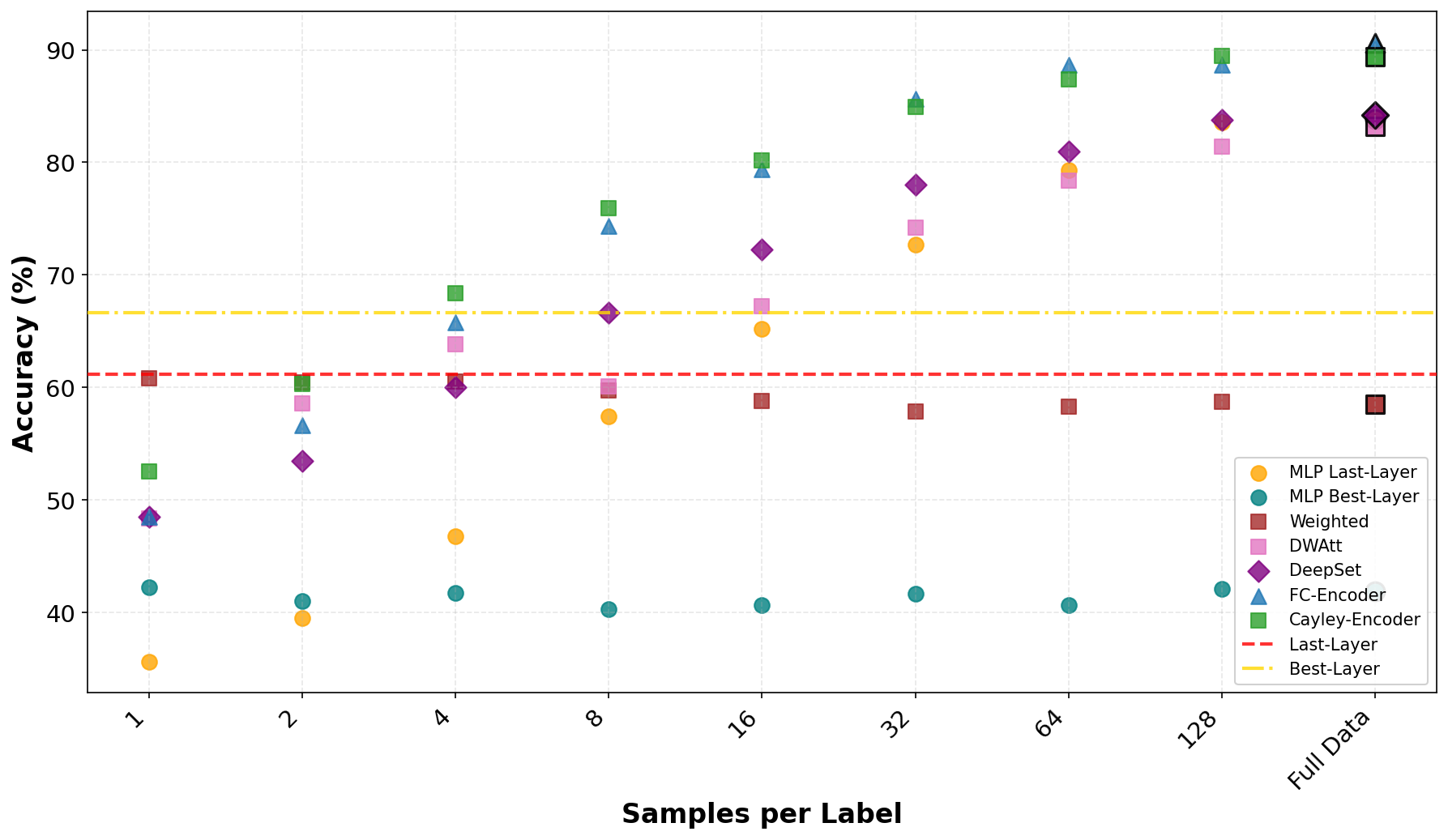}
          \caption{Banking77}
      \end{subfigure}
      \hfill
      \begin{subfigure}[b]{0.19\textwidth}
          \centering
          \includegraphics[width=\textwidth]{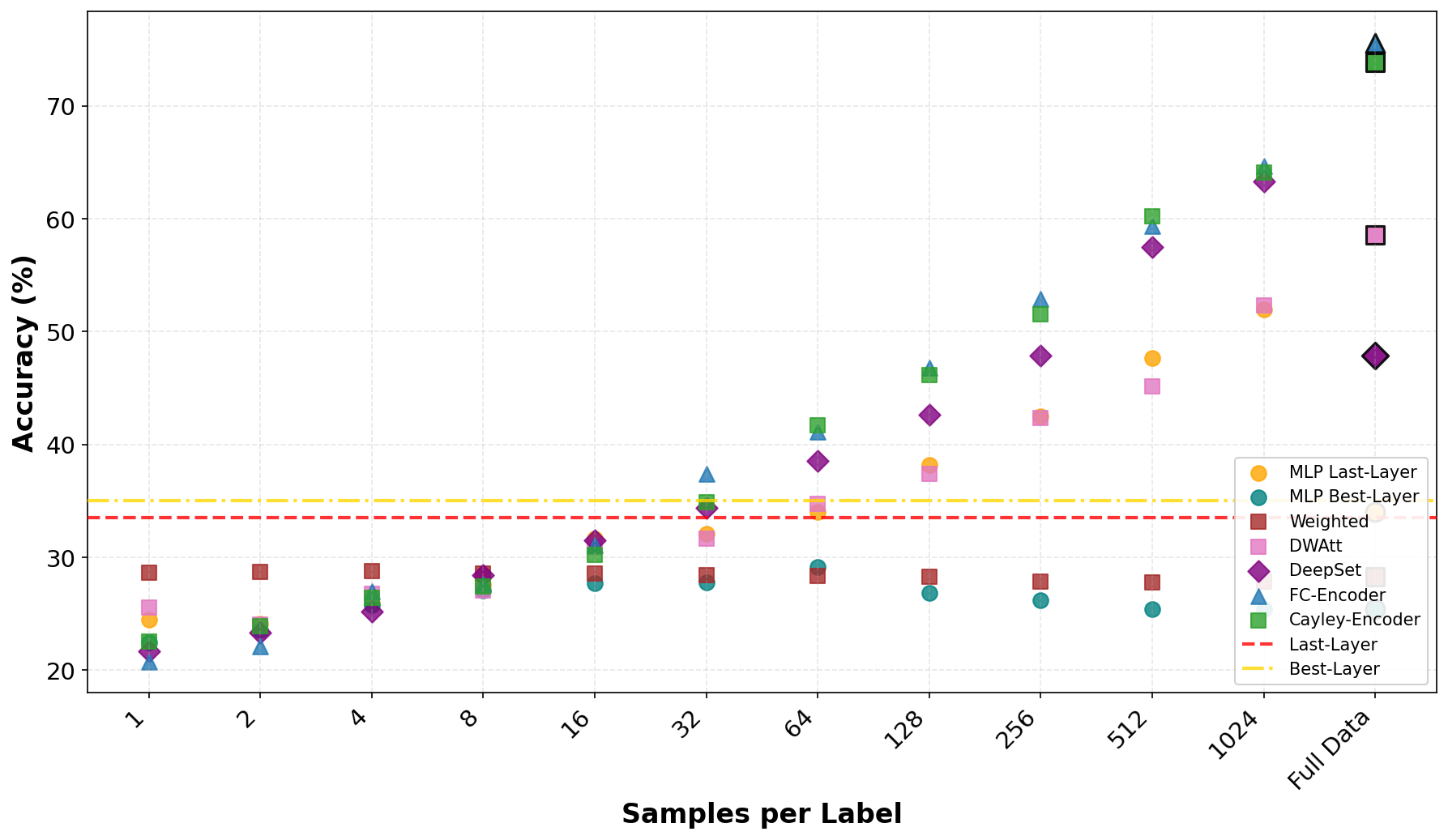}
          \caption{Emotion}
      \end{subfigure}
      \hfill
      \begin{subfigure}[b]{0.19\textwidth}
          \centering
          \includegraphics[width=\textwidth]{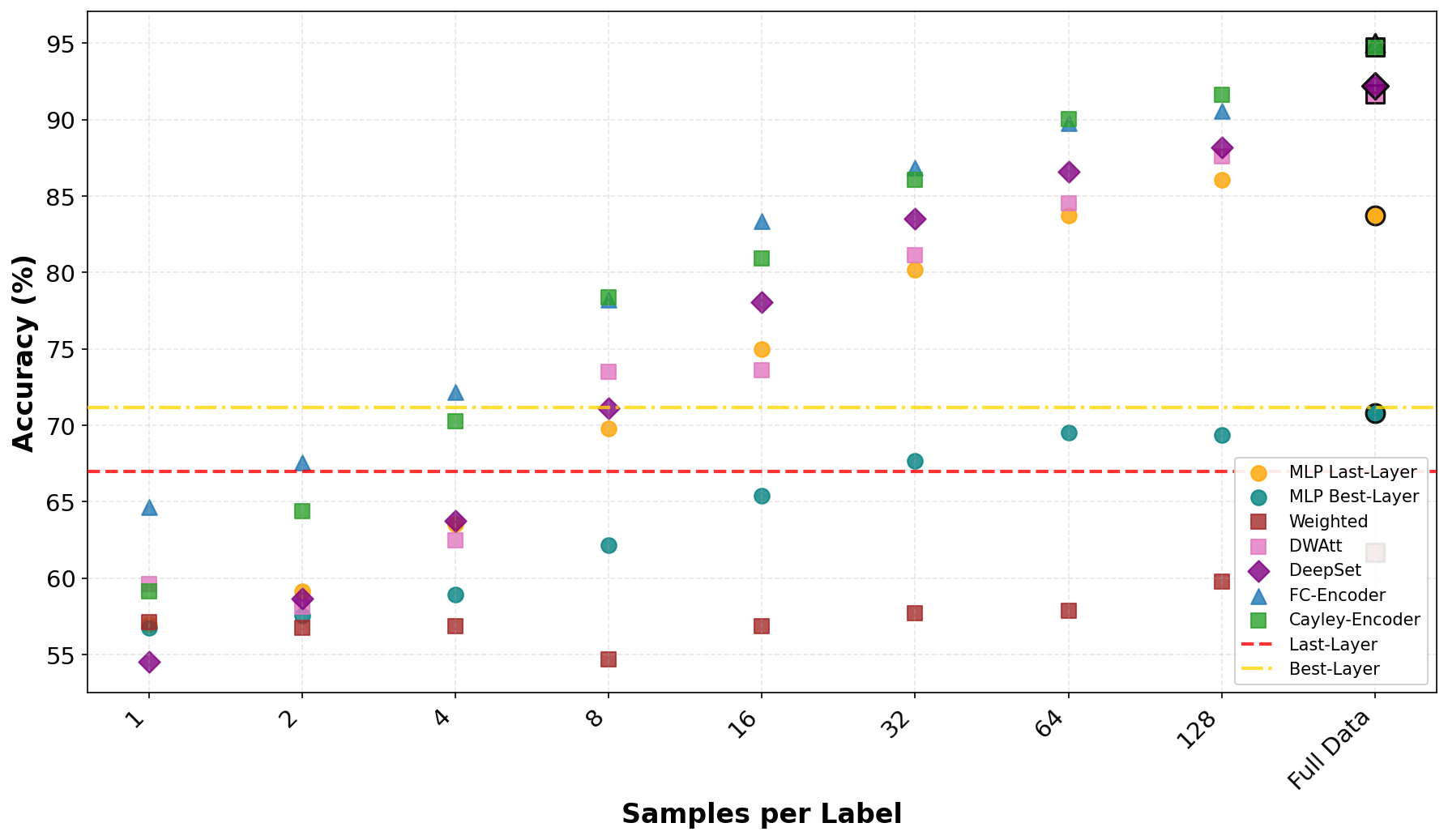}
          \caption{MTOPIntent}
      \end{subfigure}
      \hfill
      \begin{subfigure}[b]{0.19\textwidth}
          \centering
          \includegraphics[width=\textwidth]{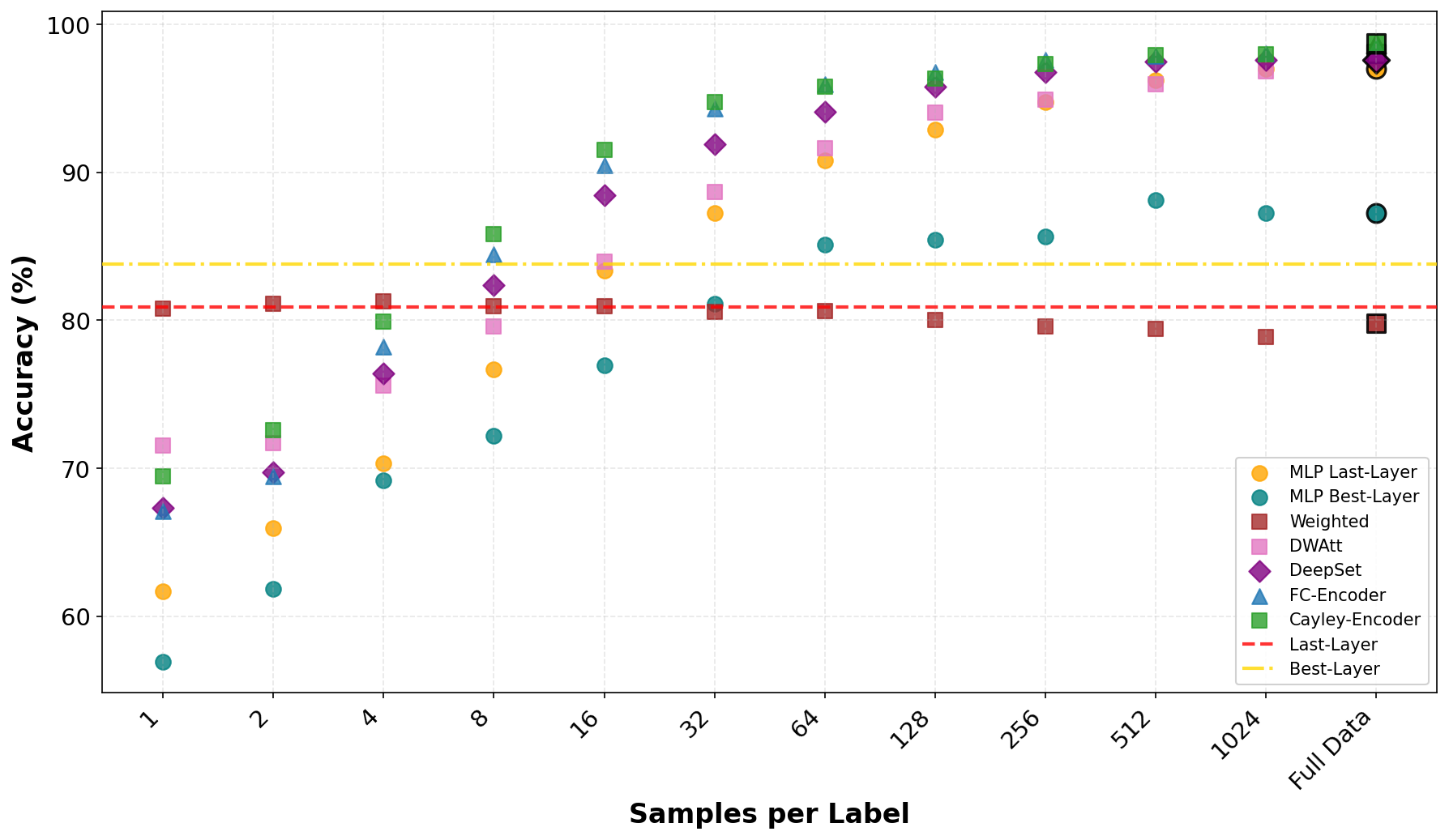}
          \caption{MTOPDomain}
      \end{subfigure}
      \hfill
      \begin{subfigure}[b]{0.19\textwidth}
          \centering
          \includegraphics[width=\textwidth]{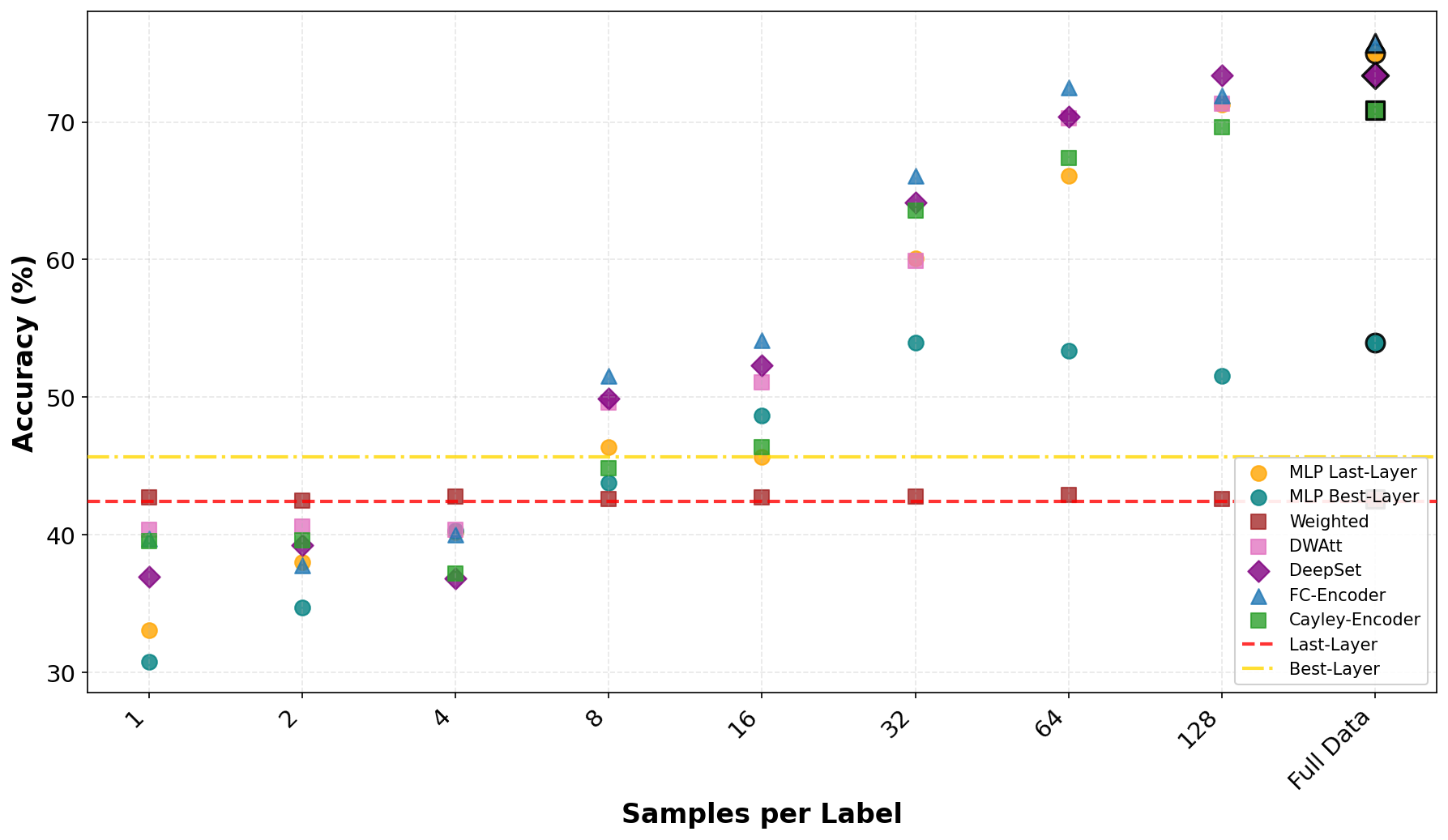}
          \caption{PoemSentiment}
      \end{subfigure}
      \caption{Few-shot learning analysis using Pythia-410M. Both encoders outperform all baselines with as few as 8 samples per label on most tasks. Cayley-Encoder consistently achieves the highest accuracy on Banking77 and the MTOP tasks across all training set sizes.}
      \label{fig:few_shot}
  \end{figure*}

We evaluate few-shot performance by restricting training to 1--1024 samples per label across all classification tasks. For Poem Sentiment, MTOPIntent, and Banking77, the available training data is limited and is fully covered at 128 samples per label. The results, presented in Figure~\ref{fig:few_shot}, show that both encoders are highly data-efficient. On Banking77, MTOPDomain, and MTOPIntent, they outperform both Last-Layer and Best-Layer baselines with as few as 8 samples per label, and Cayley-Encoder achieves the highest accuracy on all three tasks at every training set size above 8. On Banking77, 32 samples per label suffice to surpass DWAtt trained on the full dataset.  On Poem Sentiment, FC-Encoder approaches its plateau with 64 samples per label, performing comparably to other methods trained on more data. For Emotion, both encoders outperform Last-Layer and Best-Layer starting from 32 samples per label, and surpass all other baselines at larger training sizes. Emotion is the only task that exhibits consistent performance improvements as training data grows; the other tasks plateau between 128 and 512 samples per label. The overall trend suggests that both encoders extract strong representations even from very few examples, making them particularly valuable for practical applications where labeled data is scarce.

\subsection{Comparison with LoRA} 
\label{sec:lora} 
Although our encoders are designed as lightweight post-training modules that can be combined with fine-tuning approaches such as LoRA~\citep{hu2022lora}, we compare them directly against LoRA to demonstrate their standalone effectiveness. LoRA inserts trainable low-rank matrices into the LLM's attention layers, modifying the model weights themselves. Following~\citet{hu2022lora}, we adapt all attention projections in every layer and keep the feed-forward modules frozen. We searched rank $r \in \{1, 2, 3, 16\}$ via Optuna and report the best-performing rank per model and task (see Appendix~\ref{app:lora} for details). Table~\ref{tab:lora_combined} presents the results for Pythia-410M, Gemma2-2B, and Llama3-8B. For classification, at least one of our encoders outperforms LoRA in 13 out of 15 configurations. For STS, Cayley-Encoder outperforms LoRA in 21 out of 24 configurations. Notably, these gains are achieved while keeping the LLM entirely frozen and using 6--12$\times$ fewer trainable parameters than LoRA at rank 16 (Table~\ref{tab:lora-param-counts}). This suggests that structured aggregation over existing layer representations can be more effective than direct weight adaptation, while being substantially more parameter-efficient.

 \begin{table*}[!htbp]
\centering
\caption{FC-Encoder and Cayley-Encoder vs.\ LoRA on classification tasks, measured by accuracy, and STS tasks, measured by Spearman correlation $\times 100$. At least one of our encoders outperforms LoRA in 13 out of 15 classification configurations, and Cayley-Encoder outperforms LoRA in 21 out of 24 STS configurations, while keeping the LLM frozen. We highlight in bold the best result per task.}
\label{tab:lora_combined}
\setlength{\tabcolsep}{2.2pt} 
\renewcommand{\arraystretch}{0.92} 
\small
\resizebox{\textwidth}{!}{%
\begin{tabular}{ll ccccc cccccccc}
\toprule
\cmidrule(lr){3-7} \cmidrule(lr){8-15}
Base Model & Method & Banking77 & Emotion & MTOPDomain & MTOPIntent & PoemSentiment & STSBench. & STS12 & STS13 & STS14 & STS15 & STS16 & BIOSSES & SICK-R \\
\midrule
\multirow{3}{*}{Pythia-410M}
 & FC-Encoder     & \textbf{90.65} & 75.61 & 98.68 & \textbf{95.04} & \textbf{75.77} & 54.2 & 50.1 & 53.0 & 51.8 & 63.9 & 57.1 & 58.6 & \textbf{57.1} \\
 & Cayley-Encoder & 89.43 & 73.83 & \textbf{98.77} & 94.72 & 70.87 & \textbf{55.8} & \textbf{56.0} & \textbf{60.2} & \textbf{56.7} & \textbf{67.0} & \textbf{58.6} & 56.6 & 55.3 \\
 & LoRA           & 82.96 & \textbf{91.57} & 96.64 & 88.25 & 70.96 & 54.1 & 44.6 & 51.7 & 52.3 & 64.8 & 45.8 & \textbf{64.3} & 52.0 \\
\midrule
\multirow{3}{*}{Gemma2-2B}
 & FC-Encoder     & 92.39 & \textbf{79.90} & 99.07 & 95.34 & 77.12 & \textbf{62.9} & \textbf{61.8} & 55.4 & 63.9 & \textbf{74.7} & \textbf{63.0} & \textbf{66.0} & 60.1 \\
 & Cayley-Encoder & \textbf{92.58} & 69.60 & \textbf{99.16} & \textbf{96.43} & \textbf{83.27} & 61.6 & 60.4 & \textbf{64.0} & \textbf{64.6} & 73.9 & 62.7 & 64.1 & \textbf{60.3} \\
 & LoRA           & 56.77 & 70.89 & 89.65 & 80.83 & 61.92 & 58.7 & 49.1 & 57.8 & 60.4 & 69.9 & 51.8 & 58.8 & 54.2 \\
\midrule
\multirow{3}{*}{Llama3-8B}
 & FC-Encoder     & 92.38 & 71.64 & 98.99 & 96.19 & 77.98 & 59.3 & 55.2 & 53.5 & 59.9 & 73.7 & 57.2 & 62.8 & 57.5 \\
 & Cayley-Encoder & \textbf{92.85} & 73.43 & \textbf{99.03} & \textbf{96.46} & \textbf{79.04} & 63.1 & \textbf{65.3} & \textbf{63.5} & \textbf{70.2} & \textbf{77.0} & \textbf{63.1} & 55.3 & \textbf{60.7} \\
 & LoRA           & 90.82 & \textbf{91.98} & 98.59 & 94.44 & 76.15 & \textbf{65.3} & 47.8 & 56.5 & 61.6 & 72.8 & 54.7 & \textbf{69.6} & 58.9 \\
\bottomrule
\end{tabular}}
\end{table*}

\needspace{4\baselineskip}
\subsection{Token-Level Inter-Layer Aggregation}
\label{sec:token_level}
  \begin{wraptable}{r}{0.4\textwidth} \centering \footnotesize \setlength{\tabcolsep}{3pt} \caption{Token-level aggregation with Gemma2-2B. Cayley-Encoder uses 12--34$\times$ less GPU memory while achieving comparable or higher accuracy. OOM: out of memory.} \label{tab:all-tokens-gemma2} \tiny \begin{tabular}{ll|ccccc} \toprule & & \textbf{Bank.} & \textbf{Emot.} & \textbf{MT-D} & \textbf{MT-I} & \textbf{Poem} \\ \midrule \multirow{2}{*}{Acc.} & Cayley & \textbf{92.6} & 89.7 & 99.1 & \textbf{96.3} & \textbf{86.4} \\ & FC     & OOM & \textbf{90.4} & \textbf{99.2} & 93.2 & 80.2 \\ \midrule \multirow{2}{*}{GPU (MB)} & Cayley & \textbf{269} & \textbf{241} & \textbf{141} & \textbf{97} & \textbf{77} \\ & FC     & OOM & 8260 & 2978 & 1583 & 923 \\ \bottomrule \end{tabular}  \end{wraptable}
In all experiments above, each layer is represented by a single mean-pooled vector over tokens, resulting in a graph with $L$ nodes (one per layer). A natural extension is to operate at the token level, constructing a graph with $L \times T$ nodes, one per token per layer, allowing the GNN to model interactions across both tokens and layers simultaneously. This setting is more expressive but substantially increases the graph size, making the choice of graph topology critical for scalability. We evaluate this extension using Gemma2-2B on all classification tasks. Table~\ref{tab:all-tokens-gemma2} presents the results. FC-Encoder exceeds GPU memory on Banking77, while Cayley-Encoder handles all tasks, using 12--34$\times$ less GPU memory. This demonstrates the practical advantage of the Cayley graph's sparse $O(n)$ edge complexity over the $O(n^2)$ cost of a fully connected graph as sequences grow longer.  Beyond efficiency, token-level aggregation improves accuracy. Compared with the mean-pooled setting (Table~\ref{tab:results}), Cayley-Encoder with token-level representations achieves higher or comparable accuracy on all classification tasks for Gemma2-2B. Emotion improves from 69.60 to 89.70 and Poem Sentiment from 83.27 to 86.44. This indicates that jointly modeling information across tokens and layers captures complementary signals that are lost by mean-pooling, and suggests that token-level inter-layer aggregation is a promising direction, particularly when paired with a scalable graph topology such as the Cayley graph.

\subsection{Understanding Inter-Layer Communication} 

\begin{wraptable}{r}{0.4\textwidth} \vspace{-12pt} \centering \footnotesize \setlength{\tabcolsep}{2pt} \caption{Robustness to layer-to-node assignment (mean $\pm$ std over 10 random permutations).} \label{tab:shuffled} \tiny\begin{tabular}{lccccc} \toprule & Bank. & Emot. & MT-D & MT-I & Poem \\ \midrule Pythia & 90.1\tiny{$\pm$.3} & 74.9\tiny{$\pm$.8} & 98.7\tiny{$\pm$.1} & 94.6\tiny{$\pm$.4} & 71.3\tiny{$\pm$1.8} \\ Gemma & 92.4\tiny{$\pm$.2} & 73.1\tiny{$\pm$1.5} & 99.1\tiny{$\pm$.1} & 96.2\tiny{$\pm$.1} & 82.3\tiny{$\pm$.5} \\ Llama & 92.9\tiny{$\pm$.1} & 72.8\tiny{$\pm$1.3} & 98.9\tiny{$\pm$.1} & 96.4\tiny{$\pm$.2} & 79.2\tiny{$\pm$1.3} \\ \bottomrule \end{tabular} \vspace{-10pt} \end{wraptable} 
To understand how different layers contribute to the final prediction in Cayley-Encoder, and what aspects of their communication are crucial, we conduct five complementary analyses.

\paragraph{Layer-to-node assignment does not matter.} Since the Cayley graph may contain virtual nodes (when $|V_n| > L$), the specific layer-to-node assignment determines which layers are directly connected and which communicate only through virtual nodes. If the identity of communicating layers mattered, this assignment would significantly affect performance. To test this, we fix the model weights and hyperparameters for Cayley-Encoder and re-evaluate with 10 independent random layer-to-node permutations across all classification tasks using Pythia-410M, Gemma2-2B, and Llama3-8B. Table~\ref{tab:shuffled} shows that accuracy remains stable with minimal variance (standard deviations below 2 percentage points in all cases), indicating that the assignment has little effect.
\paragraph{Virtual nodes do not dilute the representation.} For small LLMs, virtual nodes can outnumber the layers. For example, Pythia-14M and Pythia-70M map 7 layer representations to a 24-node Cayley graph, so 71\% of the nodes are virtual, and Cayley-Encoder still outperforms all baselines (Figure~\ref{fig:model_scaling}). To test this directly, we train Cayley-Encoder with Gemma2-2B on graphs larger than required, increasing the fraction of virtual nodes from 44\% ($n=4$, 48 nodes) to 78\% ($n=5$, 120 nodes) and 81\% ($n=6$, 144 nodes). STS performance does not decrease up to 78\% and decreases only slightly at 81\%, with 5 out of 8 tasks unaffected (Appendix~\ref{app:dilution}).

\begin{wraptable}{r}{0.4\textwidth} \vspace{-12pt} \centering \footnotesize \setlength{\tabcolsep}{2pt} \caption{Effect of adding learnable layer-position encodings. Layer identity information degrades performance in most cases.} \label{tab:pe_ablation} \tiny \begin{tabular}{lccccc} \toprule & Bank. & Emot. & MT-D & MT-I & Poem \\ \midrule Cayley      & 89.4 & 73.8 & 98.8 & 94.7 & 70.9 \\ Cayley (PE) & 90.3 & 75.5 & 98.7 & 94.6 & 69.2 \\ \midrule FC          & 90.7 & 75.6 & 98.7 & 95.0 & 75.8 \\ FC (PE)     & 86.3 & 72.1 & 98.4 & 94.1 & 68.4 \\ \bottomrule \end{tabular} \vspace{-10pt} \end{wraptable}
\paragraph{Layer identity is not a useful signal.} We augment each node in the layer graph with a learnable positional encoding indexed by its original depth in the LLM, added to the node features prior to message passing. This provides the GNN with explicit information about which LLM layer each node represents. We train and evaluate on all classification tasks using Pythia-410M. Table~\ref{tab:pe_ablation} shows that adding positional encodings degrades performance in most cases, for both Cayley-Encoder and FC-Encoder. This suggests that layer identity is not a useful signal for the aggregation.


\begin{wraptable}{r}{0.4\textwidth}
\centering
\footnotesize
\setlength{\tabcolsep}{2pt}
\caption{Effect of the Cayley graph degree (Qwen3-8B). The degree-4 graph used throughout the paper achieves the best result on 4 of 5 tasks and the best average.}
\label{tab:cay-deg}
\tiny
\begin{tabular}{lcccccc}
\toprule
 & Bank. & Emot. & MT-D & MT-I & Poem & Avg \\
\midrule
Cayley (deg 4) & \textbf{92.34} & \textbf{75.70} & \textbf{98.99} & \textbf{95.99} & 85.38 & \textbf{89.68} \\
Cayley (deg 6) & 92.01 & 75.23 & 98.96 & 95.77 & 78.75 & 88.14 \\
Cayley (deg 8) & 92.31 & 66.28 & 98.74 & 94.76 & \textbf{87.21} & 87.86 \\
\bottomrule
\end{tabular}
\vspace{-10pt}
\end{wraptable}
\paragraph{Cayley Degree 4 vs 6 and 8.}
Our Cayley graphs are 4-regular, since they are generated by the Margulis generating set (Appendix~\ref{app:cayley}). To evaluate the effect of the degree, we construct Cayley graphs over $SL(2,\mathbb{Z}_n)$ of degree 6 and 8 by extending the generating set with additional generators and their inverses (Appendix~\ref{app:cayley_degree}). We evaluate them using Qwen3-8B on all classification tasks. Table~\ref{tab:cay-deg} shows that the degree-6 and degree-8 graphs do not improve over the degree-4 graph, which achieves the best result on 4 out of 5 tasks and the highest average. We therefore use the degree-4 graph, the sparsest of the three, throughout the paper.

\begin{figure*}[!htbp]
      \centering                                                                            
      \includegraphics[width=\textwidth]{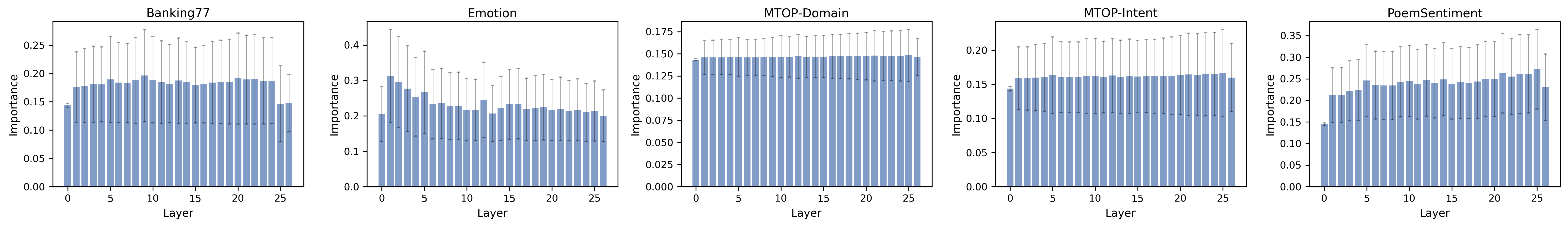}
      \caption{Per-layer contribution to the final prediction for Cayley-Encoder across classification tasks using Gemma2-2B. Estimated via GNNExplainer and averaged over all test instances. Error bars show standard deviation. All layers contribute roughly equally.}
      \label{fig:layer_importance}
        \vspace{-1.0em}
  \end{figure*}

\paragraph{All layers contribute.} We apply GNNExplainer~\citep{ying2019gnnexplainer}, a post-hoc attribution method for GNNs, to Cayley-Encoder trained on each classification task with Gemma2-2B. For each test instance, GNNExplainer identifies which nodes in the layer graph the model relied on most for its prediction. Figure~\ref{fig:layer_importance} shows the mean node importance per layer, averaged over all test instances across all five tasks. All layers contribute roughly equally, with no single layer dominating. This suggests that Cayley-Encoder aggregates complementary information from across the full depth of the LLM rather than collapsing to a single layer.

Together, these results suggest that the primary driver of Cayley-Encoder's effectiveness is enabling all layers to communicate through an expressive graph topology, rather than the specific identity or ordering of the layers. The benefit lies in structured inter-layer communication itself, not in privileging particular layers or positions.
This is consistent with the role of the graph in our method. We do not claim that any specific connectivity between layers carries task information. Rather, any aggregation over layers imposes some structure, and since this structure is independent of the target variable, we choose one that reduces the risk of overfitting to it while ensuring efficient communication between all layers. In line with this, replacing the Cayley graph with a random 4-regular graph, which does not guarantee these properties, degrades performance (Appendix~\ref{app:graph_ablation}).

\section{Conclusion}\label{sec:conclusion}

We introduced Cayley-Encoder, a lightweight method for aggregating representations from all internal layers of a frozen LLM through a structured graph topology. Across 13 diverse tasks and 13 LLMs, Cayley-Encoder consistently outperformed all baselines by large margins, including attention-based aggregation, best-layer selection, and LoRA fine-tuning, while adding at most 0.1\% parameters. These gains persist in few-shot regimes and across LLM sizes from 14M to 14B parameters. Our analysis reveals that the benefit stems from enabling effective communication across all layers, rather than from the identity or ordering of individual layers. This suggests that the internal depth of LLMs contains complementary task-relevant signals that standard last-layer extraction fails to exploit, and that structured inter-layer aggregation provides a simple and efficient way to unlock them.

Several promising directions emerge. Scaling to larger architectures (e.g., 70B+ parameters) and generative settings remains to be explored. Additionally, integrating structured inter-layer communication directly into LLM pre-training, rather than as a post-training module, could yield models with improved information flow across depth.

\begin{ack}
  This study was supported in part by a fellowship from the Edmond J. Safra Center for Bioinformatics at Tel Aviv University.
  \end{ack}

\clearpage
\bibliographystyle{plainnat}
\bibliography{ilse}

@article{hu2022lora,
  title={Lora: Low-rank adaptation of large language models.},
  author={Hu, Edward J and Shen, Yelong and Wallis, Phillip and Allen-Zhu, Zeyuan and Li, Yuanzhi and Wang, Shean and Wang, Liang and Chen, Weizhu and others},
  journal={Iclr},
  volume={1},
  number={2},
  pages={3},
  year={2022}
}

@inproceedings{vaswani2017attention,
 author = {Vaswani, Ashish and Shazeer, Noam and Parmar, Niki and Uszkoreit, Jakob and Jones, Llion and Gomez, Aidan N and Kaiser, \L ukasz and Polosukhin, Illia},
 booktitle = {Advances in Neural Information Processing Systems},
 pages = {},
 publisher = {Curran Associates, Inc.},
 title = {Attention is All you Need},
 url = {https://proceedings.neurips.cc/paper_files/paper/2017/file/3f5ee243547dee91fbd053c1c4a845aa-Paper.pdf},
 volume = {30},
 year = {2017}
}

@inproceedings{zaheer2017deep,
 author = {Zaheer, Manzil and Kottur, Satwik and Ravanbakhsh, Siamak and Poczos, Barnabas and Salakhutdinov, Russ R and Smola, Alexander J},
 booktitle = {Advances in Neural Information Processing Systems}, 
 pages = {},
 title = {Deep Sets},
 url = {https://proceedings.neurips.cc/paper_files/paper/2017/file/f22e4747da1aa27e363d86d40ff442fe-Paper.pdf},
 volume = {30},
 year = {2017}
}

@inproceedings{devlin2019bert,
    title = "{BERT}: Pre-training of Deep Bidirectional Transformers for Language Understanding",
    author = "Devlin, Jacob  and
      Chang, Ming-Wei  and
      Lee, Kenton  and
      Toutanova, Kristina",    
    booktitle = "Proceedings of the 2019 Conference of the North {A}merican Chapter of the Association for Computational Linguistics: Human Language Technologies, Volume 1 (Long and Short Papers)",
    month = jun,
    year = "2019",
    url = "https://aclanthology.org/N19-1423/",
    doi = "10.18653/v1/N19-1423",
    pages = "4171--4186"
}

@inproceedings{brown2020language,
 author = {Brown, Tom and Mann, Benjamin and Ryder, Nick and Subbiah, Melanie and Kaplan, Jared D and Dhariwal, Prafulla and Neelakantan, Arvind and Shyam, Pranav and Sastry, Girish and Askell, Amanda and Agarwal, Sandhini and Herbert-Voss, Ariel and Krueger, Gretchen and Henighan, Tom and Child, Rewon and Ramesh, Aditya and Ziegler, Daniel and Wu, Jeffrey and Winter, Clemens and Hesse, Chris and Chen, Mark and Sigler, Eric and Litwin, Mateusz and Gray, Scott and Chess, Benjamin and Clark, Jack and Berner, Christopher and McCandlish, Sam and Radford, Alec and Sutskever, Ilya and Amodei, Dario},
 booktitle = {Advances in Neural Information Processing Systems},
 editor = {H. Larochelle and M. Ranzato and R. Hadsell and M.F. Balcan and H. Lin},
 pages = {1877--1901},
 publisher = {Curran Associates, Inc.},
 title = {Language Models are Few-Shot Learners},
 url = {https://proceedings.neurips.cc/paper_files/paper/2020/file/1457c0d6bfcb4967418bfb8ac142f64a-Paper.pdf},
 volume = {33},
 year = {2020}
}

@ARTICLE{10670406,
  author={Gkarmpounis, Georgios and Vranis, Christos and Vretos, Nicholas and Daras, Petros},
  journal={IEEE Access}, 
  title={Survey on Graph Neural Networks}, 
  year={2024},
  volume={12},
  number={},
  pages={128816-128832},
  doi={10.1109/ACCESS.2024.3456913}}

@inproceedings{reimers2019sentence,                                                                                                                                                                                                                                        
    title={Sentence-BERT: Sentence Embeddings using Siamese BERT-Networks},
    author={Reimers, Nils and Gurevych, Iryna},                                                                                                                                                                                                                              
    booktitle={Proceedings of the 2019 Conference on Empirical Methods in Natural Language Processing (EMNLP)},                                                                                                                                                              
    year={2019}                                                                                                                                                                                                                                                              
  }

@inproceedings{fan2019reducing,                                                                                                                                                                                                                                            
    title={Reducing Transformer Depth on Demand with Structured Dropout},                                                                                                                                                                                                    
    author={Fan, Angela and Grave, Edouard and Joulin, Armand},                                                                                                                                                                                                              
    booktitle={International Conference on Learning Representations (ICLR)},                                                                                                                                                                                                 
    year={2020}                                                                                                                                                                                                                                                              
  }

@inproceedings{tenney2019bert,
    title = "{BERT} Rediscovers the Classical {NLP} Pipeline",
    author = "Tenney, Ian  and
      Das, Dipanjan  and
      Pavlick, Ellie",    
    booktitle = "Proceedings of the 57th Annual Meeting of the Association for Computational Linguistics",
    month = jul,
    year = "2019",
    url = "https://aclanthology.org/P19-1452/",
    doi = "10.18653/v1/P19-1452",
    pages = "4593--4601"
}

@inproceedings{jawahar2019does,
    title = "What Does {BERT} Learn about the Structure of Language?",
    author = "Jawahar, Ganesh  and
      Sagot, Beno{\^i}t  and
      Seddah, Djam{\'e}",
    editor = "Korhonen, Anna  and
      Traum, David  and
      M{\`a}rquez, Llu{\'i}s",
    booktitle = "Proceedings of the 57th Annual Meeting of the Association for Computational Linguistics",
    month = jul,
    year = "2019",
    address = "Florence, Italy",
    publisher = "Association for Computational Linguistics",
    url = "https://aclanthology.org/P19-1356/",
    doi = "10.18653/v1/P19-1356",
    pages = "3651--3657"
}

@misc{skean2025layer,
      title={Layer by Layer: Uncovering Hidden Representations in Language Models}, 
      author={Oscar Skean and Md Rifat Arefin and Dan Zhao and Niket Patel and Jalal Naghiyev and Yann LeCun and Ravid Shwartz-Ziv},
      year={2025},
      eprint={2502.02013},
      archivePrefix={arXiv},
      primaryClass={cs.LG},
      url={https://arxiv.org/abs/2502.02013}, 
}

@inproceedings{peters-etal-2018-deep,
    title = "Deep Contextualized Word Representations",
    author = "Peters, Matthew E.  and
      Neumann, Mark  and
      Iyyer, Mohit  and
      Gardner, Matt  and
      Clark, Christopher  and
      Lee, Kenton  and
      Zettlemoyer, Luke",    
    booktitle = "Proceedings of the 2018 Conference of the North {A}merican Chapter of the Association for Computational Linguistics: Human Language Technologies, Volume 1 (Long Papers)",
    month = jun,
    year = "2018",    
    url = "https://aclanthology.org/N18-1202/",
    doi = "10.18653/v1/N18-1202",
    pages = "2227--2237"    
}

@inproceedings{wallat2021bertnesia,
    title = "{BERT}nesia: Investigating the capture and forgetting of knowledge in {BERT}",
    author = "Wallat, Jonas  and
      Singh, Jaspreet  and
      Anand, Avishek",    
    booktitle = "Proceedings of the Third BlackboxNLP Workshop on Analyzing and Interpreting Neural Networks for NLP",
    month = nov,
    year = "2020",  
    url = "https://aclanthology.org/2020.blackboxnlp-1.17/",
    doi = "10.18653/v1/2020.blackboxnlp-1.17",
    pages = "174--183"
}

@inproceedings{elnokrashy2024depth,
    title = "Depth-Wise Attention ({DWA}tt): A Layer Fusion Method for Data-Efficient Classification",
    author = "ElNokrashy, Muhammad  and
      AlKhamissi, Badr  and
      Diab, Mona",    
    booktitle = "Proceedings of the 2024 Joint International Conference on Computational Linguistics, Language Resources and Evaluation (LREC-COLING 2024)",
    month = may,
    year = "2024",    
    url = "https://aclanthology.org/2024.lrec-main.417/",
    pages = "4665--4674"
}

@misc{pagliardini2024denseformer,
      title={DenseFormer: Enhancing Information Flow in Transformers via Depth Weighted Averaging}, 
      author={Matteo Pagliardini and Amirkeivan Mohtashami and Francois Fleuret and Martin Jaggi},
      year={2024},
      eprint={2402.02622},
      archivePrefix={arXiv},
      primaryClass={cs.CL},
      url={https://arxiv.org/abs/2402.02622}, 
}

@misc{kipf2016semi,
      title={Semi-Supervised Classification with Graph Convolutional Networks}, 
      author={Thomas N. Kipf and Max Welling},
      year={2017},
      eprint={1609.02907},
      archivePrefix={arXiv},
      primaryClass={cs.LG},
      url={https://arxiv.org/abs/1609.02907}, 
}

@misc{xu2018powerful,
      title={How Powerful are Graph Neural Networks?}, 
      author={Keyulu Xu and Weihua Hu and Jure Leskovec and Stefanie Jegelka},
      year={2019},
      eprint={1810.00826},
      archivePrefix={arXiv},
      primaryClass={cs.LG},
      url={https://arxiv.org/abs/1810.00826}, 
}

@InProceedings{wilson2025cayley,
  title = 	 {Cayley Graph Propagation},
  author =       {Wilson, JJ and Bechler-Speicher, Maya and Veli{\v c}kovi{\' c}, Petar},
  booktitle = 	 {Proceedings of the Third Learning on Graphs Conference},
  pages = 	 {8:1--8:20},
  year = 	 {2025},
  volume = 	 {269},
  series = 	 {Proceedings of Machine Learning Research},
  month = 	 {26--29 Nov},  
  url = 	 {https://proceedings.mlr.press/v269/wilson25a.html},  
}

@InProceedings{deac2022expander,
  title = 	 {Expander Graph Propagation},
  author =       {Deac, Andreea and Lackenby, Marc and Veli{\v c}kovi{\' c}, Petar},
  booktitle = 	 {Proceedings of the First Learning on Graphs Conference},
  pages = 	 {38:1--38:18},
  year = 	 {2022},
  volume = 	 {198},
  series = 	 {Proceedings of Machine Learning Research},
  month = 	 {09--12 Dec},
  url = 	 {https://proceedings.mlr.press/v198/deac22a.html},
}

@misc{alon2020bottleneck,
      title={On the Bottleneck of Graph Neural Networks and its Practical Implications}, 
      author={Uri Alon and Eran Yahav},
      year={2021},
      eprint={2006.05205},
      archivePrefix={arXiv},
      primaryClass={cs.LG},
      url={https://arxiv.org/abs/2006.05205}, 
}

@inproceedings{muennighoff2023mteb,
  title={Mteb: Massive text embedding benchmark},
  author={Muennighoff, Niklas and Tazi, Nouamane and Magne, Lo{\"\i}c and Reimers, Nils},
  booktitle={Proceedings of the 17th Conference of the European Chapter of the Association for Computational Linguistics},
  pages={2014--2037},
  year={2023}
}

@article{kingma2014adam,
  title={Adam: A method for stochastic optimization},
  author={Kingma, Diederik P},
  journal={arXiv preprint arXiv:1412.6980},
  year={2014}
}

@inproceedings{akiba2019optuna,
  title={Optuna: A next-generation hyperparameter optimization framework},
  author={Akiba, Takuya and Sano, Shotaro and Yanase, Toshihiko and Ohta, Takeru and Koyama, Masanori},
  booktitle={Proceedings of the 25th ACM SIGKDD international conference on knowledge discovery \& data mining},
  pages={2623--2631},
  year={2019}
}

@inproceedings{clark-etal-2019-bert,
    title = "What Does {BERT} Look at? An Analysis of {BERT}{'}s Attention",
    author = "Clark, Kevin  and
      Khandelwal, Urvashi  and
      Levy, Omer  and
      Manning, Christopher D.",
    editor = "Linzen, Tal  and
      Chrupa{\l}a, Grzegorz  and
      Belinkov, Yonatan  and
      Hupkes, Dieuwke",
    booktitle = "Proceedings of the 2019 ACL Workshop BlackboxNLP: Analyzing and Interpreting Neural Networks for NLP",
    month = aug,
    year = "2019",
    address = "Florence, Italy",
    publisher = "Association for Computational Linguistics",
    url = "https://aclanthology.org/W19-4828/",
    doi = "10.18653/v1/W19-4828",
    pages = "276--286"
}

@article{xiao2025muddformer,
  title={Muddformer: Breaking residual bottlenecks in transformers via multiway dynamic dense connections},
  author={Xiao, Da and Meng, Qingye and Li, Shengping and Yuan, Xingyuan},
  journal={arXiv preprint arXiv:2502.12170},
  year={2025}
}

@inproceedings{biderman2023pythia,
  title={Pythia: A suite for analyzing large language models across training and scaling},
  author={Biderman, Stella and Schoelkopf, Hailey and Anthony, Quentin Gregory and Bradley, Herbie and O’Brien, Kyle and Hallahan, Eric and Khan, Mohammad Aflah and Purohit, Shivanshu and Prashanth, USVSN Sai and Raff, Edward and others},
  booktitle={International Conference on Machine Learning},
  pages={2397--2430},
  year={2023},
  organization={PMLR}
}

@article{llama3modelcard,
  title={Llama 3 Model Card},
  author={AI@Meta},
  year={2024},
  url = {https://github.com/meta-llama/llama3/blob/main/MODEL_CARD.md}
}

@inproceedings{gilmer2017neural,
  title={Neural message passing for quantum chemistry},
  author={Gilmer, Justin and Schoenholz, Samuel S and Riley, Patrick F and Vinyals, Oriol and Dahl, George E},
  booktitle={International conference on machine learning},
  pages={1263--1272},
  year={2017},
  organization={Pmlr}
}

@inproceedings{bechler2024graph,
  title={Graph neural networks use graphs when they shouldn't},
  author={Bechler-Speicher, Maya and Amos, Ido and Gilad-Bachrach, Ran and Globerson, Amir},
  booktitle={Forty-first International Conference on Machine Learning},
  year={2024}
}

@article{fan2024not,
  title={Not all layers of llms are necessary during inference},
  author={Fan, Siqi and Jiang, Xin and Li, Xiang and Meng, Xuying and Han, Peng and Shang, Shuo and Sun, Aixin and Wang, Yequan and Wang, Zhongyuan},
  journal={arXiv preprint arXiv:2403.02181},
  year={2024}
}

@inproceedings{li-etal-2021-mtop,
  address = {Online},
  author = {Li, Haoran  and
Arora, Abhinav  and
Chen, Shuohui  and
Gupta, Anchit  and
Gupta, Sonal  and
Mehdad, Yashar},
  booktitle = {Proceedings of the 16th Conference of the European Chapter of the Association for Computational Linguistics: Main Volume},
  doi = {10.18653/v1/2021.eacl-main.257},
  editor = {Merlo, Paola  and
Tiedemann, Jorg  and
Tsarfaty, Reut},
  month = apr,
  pages = {2950--2962},
  publisher = {Association for Computational Linguistics},
  title = {{MTOP}: A Comprehensive Multilingual Task-Oriented Semantic Parsing Benchmark},
  url = {https://aclanthology.org/2021.eacl-main.257},
  year = {2021},
}

@inproceedings{saravia-etal-2018-carer,
  address = {Brussels, Belgium},
  author = {Saravia, Elvis  and
Liu, Hsien-Chi Toby  and
Huang, Yen-Hao  and
Wu, Junlin  and
Chen, Yi-Shin},
  booktitle = {Proceedings of the 2018 Conference on Empirical Methods in Natural Language Processing},
  doi = {10.18653/v1/D18-1404},
  editor = {Riloff, Ellen  and
Chiang, David  and
Hockenmaier, Julia  and
Tsujii, Jun{'}ichi},
  month = oct # {-} # nov,
  pages = {3687--3697},
  publisher = {Association for Computational Linguistics},
  title = {{CARER}: Contextualized Affect Representations for Emotion Recognition},
  url = {https://aclanthology.org/D18-1404},
  year = {2018},
}

@inproceedings{casanueva-etal-2020-efficient,
    title = "Efficient Intent Detection with Dual Sentence Encoders",
    author = "Casanueva, I{\~n}igo  and
      Tem{\v{c}}inas, Tadas  and
      Gerz, Daniela  and
      Henderson, Matthew  and
      Vuli{\'c}, Ivan",
    editor = "Wen, Tsung-Hsien  and
      Celikyilmaz, Asli  and
      Yu, Zhou  and
      Papangelis, Alexandros  and
      Eric, Mihail  and
      Kumar, Anuj  and
      Casanueva, I{\~n}igo  and
      Shah, Rushin",
    booktitle = "Proceedings of the 2nd Workshop on Natural Language Processing for Conversational AI",
    month = jul,
    year = "2020",
    address = "Online",
    publisher = "Association for Computational Linguistics",
    url = "https://aclanthology.org/2020.nlp4convai-1.5/",
    doi = "10.18653/v1/2020.nlp4convai-1.5",
    pages = "38--45"
}

@inproceedings{sheng2020investigating,
  title={Investigating societal biases in a poetry composition system},
  author={Sheng, Emily and Uthus, David C},
  booktitle={Proceedings of the Second Workshop on Gender Bias in Natural Language Processing},
  pages={93--106},
  year={2020}
}

@inproceedings{huggingface:dataset:stsb_multi_mt,
  author = {Philip May},
  title = {Machine translated multilingual STS benchmark dataset.},
  url = {https://github.com/PhilipMay/stsb-multi-mt},
  year = {2021},
}

@inproceedings{10.5555/2387636.2387697,
  address = {USA},
  author = {Agirre, Eneko and Diab, Mona and Cer, Daniel and Gonzalez-Agirre, Aitor},
  booktitle = {Proceedings of the First Joint Conference on Lexical and Computational Semantics - Volume 1: Proceedings of the Main Conference and the Shared Task, and Volume 2: Proceedings of the Sixth International Workshop on Semantic Evaluation},
  location = {Montr\'{e}al, Canada},
  numpages = {9},
  pages = {385–393},
  publisher = {Association for Computational Linguistics},
  series = {SemEval '12},
  title = {SemEval-2012 task 6: a pilot on semantic textual similarity},
  year = {2012},
}

@inproceedings{Agirre2013SEM2S,
  author = {Eneko Agirre and Daniel Matthew Cer and Mona T. Diab and Aitor Gonzalez-Agirre and Weiwei Guo},
  booktitle = {International Workshop on Semantic Evaluation},
  title = {*SEM 2013 shared task: Semantic Textual Similarity},
  url = {https://api.semanticscholar.org/CorpusID:10241043},
  year = {2013},
}

@article{10.1093/bioinformatics/btx238,
  author = {Soğancıoğlu, Gizem and Öztürk, Hakime and Özgür, Arzucan},
  doi = {10.1093/bioinformatics/btx238},
  eprint = {https://academic.oup.com/bioinformatics/article-pdf/33/14/i49/50315066/bioinformatics\_33\_14\_i49.pdf},
  issn = {1367-4803},
  journal = {Bioinformatics},
  month = {07},
  number = {14},
  pages = {i49-i58},
  title = {{BIOSSES: a semantic sentence similarity estimation system for the biomedical domain}},
  url = {https://doi.org/10.1093/bioinformatics/btx238},
  volume = {33},
  year = {2017},
}

@inproceedings{marelli-etal-2014-sick,
  address = {Reykjavik, Iceland},
  author = {Marelli, Marco  and
Menini, Stefano  and
Baroni, Marco  and
Bentivogli, Luisa  and
Bernardi, Raffaella  and
Zamparelli, Roberto},
  booktitle = {Proceedings of the Ninth International Conference on Language Resources and Evaluation ({LREC}'14)},
  editor = {Calzolari, Nicoletta  and
Choukri, Khalid  and
Declerck, Thierry  and
Loftsson, Hrafn  and
Maegaard, Bente  and
Mariani, Joseph  and
Moreno, Asuncion  and
Odijk, Jan  and
Piperidis, Stelios},
  month = may,
  pages = {216--223},
  publisher = {European Language Resources Association (ELRA)},
  title = {A {SICK} cure for the evaluation of compositional distributional semantic models},
  url = {http://www.lrec-conf.org/proceedings/lrec2014/pdf/363_Paper.pdf},
  year = {2014},
}

@article{gemma_2024,
    title={Gemma},
    url={https://www.kaggle.com/m/3301},
    DOI={10.34740/KAGGLE/M/3301},
    publisher={Kaggle},
    author={Gemma Team},
    year={2024}
}

@article{ying2019gnnexplainer,
  title={Gnnexplainer: Generating explanations for graph neural networks},
  author={Ying, Zhitao and Bourgeois, Dylan and You, Jiaxuan and Zitnik, Marinka and Leskovec, Jure},
  journal={Advances in neural information processing systems},
  volume={32},
  year={2019}
}

@article{mantri2026towards,
  title={Towards Improved Sentence Representations using Token Graphs},
  author={Mantri, Krishna Sri Ipsit and Sch{\"o}nlieb, Carola-Bibiane and L{\"a}hner, Zorah and Eliasof, Moshe},
  journal={arXiv preprint arXiv:2603.03389},
  year={2026}
}

@inproceedings{velickovic2018graph,
  title={Graph Attention Networks},
  author={Veli{\v{c}}kovi{\'c}, Petar and Cucurull, Guillem and Casanova, Arantxa and Romero, Adriana and Li{\`o}, Pietro and Bengio, Yoshua},
  booktitle={International Conference on Learning Representations},
  year={2018}
}

@article{yang2025qwen3,
  title={Qwen3 technical report},
  author={Yang, An and Li, Anfeng and Yang, Baosong and Zhang, Beichen and Hui, Binyuan and Zheng, Bo and Yu, Bowen and Gao, Chang and Huang, Chengen and Lv, Chenxu and others},
  journal={arXiv preprint arXiv:2505.09388},
  year={2025}
}

@inproceedings{agirre2014semeval,
  title={Semeval-2014 task 10: Multilingual semantic textual similarity},
  author={Agirre, Eneko and Banea, Carmen and Cardie, Claire and Cer, Daniel and Diab, Mona and Gonzalez-Agirre, Aitor and Guo, Weiwei and Mihalcea, Rada and Rigau, German and Wiebe, Janyce},
  booktitle={Proceedings of the 8th international workshop on semantic evaluation (SemEval 2014)},
  pages={81--91},
  year={2014}
}

@inproceedings{agirre2015semeval,
  title={Semeval-2015 task 2: Semantic textual similarity, english, spanish and pilot on interpretability},
  author={Agirre, Eneko and Banea, Carmen and Cardie, Claire and Cer, Daniel and Diab, Mona and Gonzalez-Agirre, Aitor and Guo, Weiwei and Lopez-Gazpio, Inigo and Maritxalar, Montse and Mihalcea, Rada and others},
  booktitle={Proceedings of the 9th international workshop on semantic evaluation (SemEval 2015)},
  pages={252--263},
  year={2015}
}

@inproceedings{agirre-etal-2016-semeval,
    title = "{S}em{E}val-2016 Task 1: Semantic Textual Similarity, Monolingual and Cross-Lingual Evaluation",
    author = "Agirre, Eneko  and
      Banea, Carmen  and
      Cer, Daniel  and
      Diab, Mona  and
      Gonzalez-Agirre, Aitor  and
      Mihalcea, Rada  and
      Rigau, German  and
      Wiebe, Janyce",
    editor = "Bethard, Steven  and
      Carpuat, Marine  and
      Cer, Daniel  and
      Jurgens, David  and
      Nakov, Preslav  and
      Zesch, Torsten",
    booktitle = "Proceedings of the 10th {I}nternational {W}orkshop on {S}emantic {E}valuation ({S}em{E}val-2016)",
    month = jun,
    year = "2016",
    address = "San Diego, California",
    publisher = "Association for Computational Linguistics",
    url = "https://aclanthology.org/S16-1081/",
    doi = "10.18653/v1/S16-1081",
    pages = "497--511"
}

\clearpage
\appendix
\onecolumn

\section{Cayley Graph Construction Details}
\label{app:cayley}

The Cayley graph over $SL(2, \mathbb{Z}_n)$ is constructed using the symmetric generating set $S = \{A, A^{-1}, B, B^{-1}\}$, where
\[
A = \begin{pmatrix} 1 & 1 \\ 0 & 1 \end{pmatrix}, \qquad
B = \begin{pmatrix} 1 & 0 \\ 1 & 1 \end{pmatrix},
\]
with all operations performed modulo $n$. These are the standard Margulis generators~\citep{deac2022expander}, producing a 4-regular graph with logarithmic diameter. Each element of $SL(2, \mathbb{Z}_n)$ corresponds to a node in the graph, and two nodes are connected if one can be obtained from the other by left-multiplication with a generator. The resulting graph is vertex-transitive, meaning all nodes are structurally equivalent.

\section{Trainable Parameter Counts}
\label{app:param-counts}

Table~\ref{tab:t-param-counts} reports the number of trainable parameters introduced by each method. All base LLM parameters remain frozen throughout. FC-Encoder and Cayley-Encoder introduce at most 0.1\% additional parameters relative to the base model, while DWAtt requires 3--5$\times$ more parameters.

\begin{table}[!htbp]
  \centering
  \footnotesize
  \renewcommand{\arraystretch}{0.95}
  \caption{Trainable parameters added by each method. All base model parameters remain frozen. Percentages show overhead relative to base model size. }
  \label{tab:t-param-counts}
  \begin{tabular*}{\textwidth}{l @{\extracolsep{\fill}} cccc}
  \toprule
  Method & Pythia-410M & Gemma2-2B & Llama3-8B & Qwen3-8B \\
  \midrule
  \textit{Frozen Parameters} & \textit{410M} & \textit{2.6B} & \textit{8B} & \textit{8B} \\
  \midrule
  Weighted & 25 & 27 & 33 & 37 \\
  MLP & 394K (0.10\%) & 721K (0.028\%) & 1.18M (0.015\%) & 1.18M (0.014\%) \\
  Deepset & 394K (0.10\%) & 721K (0.028\%) & 1.18M (0.015\%) & 1.18M (0.014\%) \\
  FC-Encoder & 395K (0.10\%) & 722K (0.028\%) & 1.18M (0.015\%) & 1.18M (0.014\%) \\
  Cayley-Encoder & 395K (0.10\%) & 722K (0.028\%) & 1.18M (0.015\%) & 1.18M (0.014\%) \\
  DWAtt & 2.0M (0.49\%) & 2.46M (0.095\%) & 3.3M (0.041\%) & 3.6M (0.045\%) \\
  \bottomrule
  \end{tabular*}
    \vspace{-1.5em}
\end{table}

\section{Implementation Details}
\label{app:impl}

We provide full hyperparameter search spaces for all methods below. All methods use the Adam optimizer~\citep{kingma2014adam} with hyperparameters selected via Optuna~\citep{akiba2019optuna} using MedianPruner (5 startup trials, 10 warmup epochs, checked every 5 epochs). The number of Optuna trials varied between 20 and 50 depending on the method.

\paragraph{FC-Encoder and Cayley-Encoder.} Number of MPNN layers $\in \{1, 2\}$, dropout $\in \{0.0, 0.1, 0.2, 0.3\}$, number of MLP layers inside GIN aggregation $\in \{1, 2\}$, learning rate $\in \{10^{-4}, 10^{-3}\}$, weight decay $\in \{10^{-4}, 10^{-3}\}$. Hidden dimension is fixed to 256 and batch size to 64.

\paragraph{MLP Last Layer.} Number of MLP layers $\in \{1, 2, 3\}$, dropout $\in \{0.0, 0.1, 0.2, 0.3\}$, learning rate $\in \{10^{-4}, 10^{-3}\}$, weight decay $\in \{10^{-4}, 10^{-3}\}$. Hidden dimension is fixed to 256.

\paragraph{Deepset.} Pre-pooling layers $\in \{0, 1\}$, post-pooling layers $\in \{0, 1, 2\}$, pooling type $\in \{\text{mean}, \text{sum}\}$, dropout $\in \{0.0, 0.1, 0.2, 0.3\}$, learning rate $\in \{10^{-4}, 10^{-3}\}$, weight decay $\in \{10^{-4}, 10^{-3}\}$. Hidden dimension is fixed to 256.

\paragraph{DWAtt.} Architecture is fixed to match the original paper: bottleneck ratio $\gamma = 0.5$, positional embedding dimension $d_{\text{pos}} = 24$, and hidden dimension 256. We search dropout $\in \{0.0, 0.1, 0.2, 0.3\}$, learning rate $\in \{10^{-4}, 10^{-3}\}$, and weight decay $\in \{10^{-4}, 10^{-3}\}$.

\paragraph{Weighted.} The architecture has no tunable structure (only $L$ scalar layer weights with softmax). We search learning rate $\in \{10^{-4}, 10^{-3}, 10^{-2}\}$ and weight decay $\in \{0, 10^{-4}, 10^{-3}\}$.

\paragraph{LoRA.} Rank $r \in \{1, 2, 3, 16\}$, alpha $\in \{8, 16, 32\}$, LoRA dropout $\in \{0.0, 0.1, 0.2\}$, learning rate $\in \{10^{-4}, 10^{-3}\}$, weight decay $\in \{10^{-4}, 10^{-3}\}$. Batch size is 32 and training runs for 20 epochs.

\begin{table*}[!htbp]
\centering
\caption{Hyperparameter search spaces. All methods use Optuna with MedianPruner (5 startup trials, 10 epochs) and 20-50 trials per study.}
\label{tab:hp-search-space}

\subfloat[Shared training hyperparameters]{%
\begin{tabular}{l c}
\toprule
\textbf{Hyperparameter} & \textbf{Search Space} \\
\midrule
Learning rate     & \{1e-4, 1e-3\} \\
Weight decay      & \{1e-4, 1e-3\} \\
Dropout           & \{0.0, 0.1, 0.2, 0.3\} \\
Batch size        & 64 \\
Epochs            & 50 (cls) / 25 (STS) \\
\bottomrule
\end{tabular}
\label{tab:hp-shared}
}
\hfill
\subfloat[Cayley-Encoder / FC-Encoder]{%
\begin{tabular}{l c}
\toprule
\textbf{Hyperparameter} & \textbf{Search Space} \\
\midrule
GNN layers        & \{1, 2\} \\
GIN MLP layers    & \{1, 2\} \\
Hidden dim        & 256 \\
Readout           & \{mean, sum\} \\
\bottomrule
\end{tabular}
\label{tab:hp-gnn}
}

\vspace{0.5em}

\subfloat[MLP]{%
\begin{tabular}{l c}
\toprule
\textbf{Hyperparameter} & \textbf{Search Space} \\
\midrule
MLP layers  & \{1, 2, 3\} \\
Hidden dim  & 256 \\
\bottomrule
\end{tabular}
\label{tab:hp-mlp}
}
\hfill
\subfloat[Deepset]{%
\begin{tabular}{l c}
\toprule
\textbf{Hyperparameter} & \textbf{Search Space} \\
\midrule
Pre-pool layers   & \{0, 1\} \\
Post-pool layers  & \{0, 1, 2\} \\
Pooling type      & \{mean, sum\} \\
Hidden dim        & 256 \\
\bottomrule
\end{tabular}
\label{tab:hp-deepset}
}
\hfill
\subfloat[DWAtt]{%
\begin{tabular}{l c}
\toprule
\textbf{Hyperparameter} & \textbf{Search Space} \\
\midrule
Bottleneck ratio  & 0.5 \\
Pos.\ embed dim   & 24 \\
Hidden dim        & 256 \\
\bottomrule
\end{tabular}
\label{tab:hp-dwatt}
}

\vspace{0.5em}

\subfloat[Weighted]{%
\begin{tabular}{l c}
\toprule
\textbf{Hyperparameter} & \textbf{Search Space} \\
\midrule
Learning rate   & \{1e-4, 1e-3, 1e-2\} \\
Weight decay    & \{0, 1e-4, 1e-3\} \\
\bottomrule
\end{tabular}
\label{tab:hp-weighted}
}
\hfill
\subfloat[LoRA]{%
\begin{tabular}{l c}
\toprule
\textbf{Hyperparameter} & \textbf{Search Space} \\
\midrule
Rank ($r$)      & \{1, 2, 3, 16\} \\
Alpha           & \{8, 16, 32\} \\
LoRA dropout    & \{0.0, 0.1, 0.2\} \\
Batch size      & 32 \\
Epochs          & 20 \\
\bottomrule
\end{tabular}
\label{tab:hp-lora}
}

\end{table*}

\section{Positional Encoding Ablation}
\label{app:pe}

We evaluate the effect of adding learnable layer-position encodings to FC-Encoder and Cayley-Encoder. Each node is assigned a learned embedding indexed by its original depth in the base LLM, added to the node features prior to message passing. We train and evaluate on both classification and STS tasks using Pythia-410M.

Table~\ref{tab:pe-ablation-combined} presents the results. For classification, adding positional encodings yields comparable performance for Cayley-Encoder but consistently degrades FC-Encoder. For STS, positional encodings degrade Cayley-Encoder while providing marginal improvements for FC-Encoder. Overall, layer-position information does not add value, and in most cases hurts performance. This supports the finding from our main analysis that effective inter-layer communication, rather than layer identity, drives the encoders' effectiveness.

\begin{table*}[ht]
\centering
\caption{Positional encoding (PE) ablation on Pythia-410M. Classification: accuracy (\%), STS: Spearman correlation ($\times 100$). \textbf{Bold}: best per column within each encoder type.}
\label{tab:pe-ablation-combined}
\resizebox{\textwidth}{!}{%
\begin{tabular}{lcccccc|cccccccc|c}
\toprule
& \multicolumn{6}{c|}{\textbf{Classification}} & \multicolumn{8}{c|}{\textbf{STS}} & \\
\cmidrule(lr){2-7} \cmidrule(lr){8-15}
Method & Banking77 & Emotion & MTOPDom. & MTOPInt. & PoemSent. & Avg & STS-B & STS12 & STS13 & STS14 & STS15 & STS16 & BIOSSES & SICK-R & Avg \\
\midrule
Cayley-Encoder      & 89.43          & 73.83          & \textbf{98.77} & \textbf{94.72} & \textbf{70.87} & 85.52          & \textbf{55.8} & \textbf{56.0} & \textbf{60.2} & \textbf{56.7} & \textbf{67.0} & \textbf{58.6} & \textbf{56.6} & \textbf{55.3} & \textbf{58.3} \\
Cayley-Encoder (PE) & \textbf{90.25} & \textbf{75.52} & 98.72          & 94.58          & 69.23          & \textbf{85.66} & 55.3          & 51.4          & 54.8          & 48.2          & 63.0          & 53.2          & 44.8          & 54.2          & 53.1          \\
\midrule
FC-Encoder          & \textbf{90.65} & \textbf{75.61} & \textbf{98.68} & \textbf{95.04} & \textbf{75.77} & \textbf{87.15} & 54.2          & 50.1          & 53.0          & 51.8          & 63.9          & \textbf{57.1} & \textbf{58.6} & \textbf{57.1} & 55.7          \\
FC-Encoder (PE)     & 86.33          & 72.07          & 98.36          & 94.10          & 68.37          & 83.85          & \textbf{55.3} & \textbf{52.5} & \textbf{55.9} & \textbf{54.2} & \textbf{65.0} & 56.3          & 56.5          & 56.6          & \textbf{56.5} \\
\bottomrule
\end{tabular}%
}
\end{table*}

\section{Additional LoRA Results}
\label{app:lora}

LoRA adapters are inserted into all attention projections in every layer of the LLM: the fused query-key-value projection and the attention output projection for Pythia, and the query, key, value, and output projections for Gemma2 and Llama3. The feed-forward modules remain frozen, as in~\citet{hu2022lora}.

Table~\ref{tab:lora-param-counts} compares the trainable parameter counts of LoRA at ranks 3 and 16 against Cayley-Encoder. At rank 16, LoRA introduces 6--12$\times$ more trainable parameters than Cayley-Encoder, depending on the model size. Despite this parameter advantage, Cayley-Encoder outperforms LoRA on the majority of tasks while keeping the LLM entirely frozen.

\begin{table}[!htbp]        
    \centering
    \footnotesize                                                                                                                         \renewcommand{\arraystretch}{0.95}
    \caption{Trainable parameters for LoRA (ranks 3 and 16) compared to Cayley-Encoder.
    Percentages show overhead relative to base model size.}
    \label{tab:lora-param-counts}
    \begin{tabular*}{\textwidth}{l @{\extracolsep{\fill}} ccc}
    \toprule
    Method & Pythia-410M & Gemma2-2B & Llama3-8B \\
    \midrule
    \textit{Frozen Parameters} & \textit{410M} & \textit{2.6B} & \textit{8B} \\
    \midrule
    LoRA ($r=3$)  & 442K (0.11\%)  & 1.20M (0.046\%) & 2.56M (0.032\%) \\
    LoRA ($r=16$) & 2.36M (0.58\%) & 6.39M (0.25\%)  & 13.6M (0.17\%)  \\
    \midrule
    Cayley-Encoder & 395K (0.10\%) & 722K (0.028\%) & 1.18M (0.015\%) \\
    \bottomrule
    \end{tabular*}
  \end{table}

\section{Cayley Graph vs.\ Random Regular Graph}
\label{app:graph_ablation}

To determine whether the performance of Cayley-Encoder stems from its algebraic structure or simply from regularity and sparsity, we replace the Cayley graph with a random 4-regular graph. The random graph matches the Cayley graph in degree, regularity, and sparsity, but does not guarantee its expansion properties. We evaluate across Pythia-410M, Gemma2-2B, and Llama3-8B on all classification and STS tasks, with hyperparameters tuned independently for each method.

Tables~\ref{tab:Cayley-Encoder-vs-rr-cls} and~\ref{tab:ablation_cayley_vs_random_sts} present the results. The Cayley graph outperforms the random 4-regular baseline on 9 out of 15 classification and 22 out of 24 STS task--model pairs, with the largest gains on Llama3-8B. This indicates that the algebraic structure of the Cayley graph, not merely its regularity or sparsity, contributes meaningfully to performance.

\begin{table}[ht]
\centering
\caption{Ablation isolating the contribution of the Cayley-Encoder graph structure. Our Cayley-Encoder uses a 4-regular topology; we ask whether its gains over FC-Encoder come from sparsity alone or from the specific Cayley-Encoder connectivity. Replacing the Cayley-Encoder graph with a random 4-regular graph in classification tasks shows comparable results. Cayley-Encoder is better in 9 out of 15 cases. \textbf{Bold}: best per column.}
\label{tab:Cayley-Encoder-vs-rr-cls}
\resizebox{\columnwidth}{!}{%
\begin{tabular}{llccccc}
\toprule
Base Model & Graph Type & Banking77 & Emotion & MTOPDomain & MTOPIntent & PoemSentiment \\
\midrule
\multirow{2}{*}{Pythia-410M}
& Random-Regular & \textbf{90.38} & \textbf{75.15} & \textbf{98.80} & \textbf{94.73} & 70.77 \\
& Cayley-Encoder         & 89.43          & 73.83          & 98.77          & 94.72          & \textbf{70.87} \\
\midrule
\multirow{2}{*}{Gemma2-2B}
& Random-Regular & 92.16          & \textbf{76.50} & 99.12          & 96.21          & 82.31 \\
& Cayley-Encoder         & \textbf{92.58} & 69.60          & \textbf{99.16} & \textbf{96.43} & \textbf{83.27} \\
\midrule
\multirow{2}{*}{Llama3-8B}
& Random-Regular & 92.23          & 72.99          & 98.90          & 96.29          & \textbf{79.81} \\
& Cayley-Encoder         & \textbf{92.85} & \textbf{73.43} & \textbf{99.03} & \textbf{96.46} & 79.04 \\
\bottomrule
\end{tabular}%
}
\end{table}

\begin{table*}[!htbp]
\centering
\caption{Ablation isolating the contribution of the Cayley-Encoder graph structure. Our Cayley-Encoder uses a 4-regular topology; we ask whether its gains over FC-Encoder come from sparsity alone or from the specific Cayley connectivity. Replacing the Cayley graph with a random 4-regular graph degrades performance on 22 of 24 task--model pairs across 3 LLMs and 8 STS benchmarks, showing that the expansion properties of the Cayley graph, not regularity alone, drive the improvement. \textbf{Bold}: best per column.}
\resizebox{\textwidth}{!}{
\begin{tabular}{llcccccccc}
\toprule
Base Model & Method & STSBenchmark & STS12 & STS13 & STS14 & STS15 & STS16 & BIOSSES & SICK-R \\
\midrule
\multirow{2}{*}{Pythia-410M}
& Random-Regular & 53.19 & 51.69 & 56.81 & 52.82 & 62.19 & 54.27 & 47.81 & \textbf{55.66} \\
& Cayley-Encoder   & \textbf{55.84} & \textbf{55.97} & \textbf{60.25} & \textbf{56.65} & \textbf{66.99} & \textbf{58.63} & \textbf{56.59} & 55.34 \\
\midrule
\multirow{2}{*}{Gemma2-2B}
& Random-Regular & 59.85 & 52.57 & 58.25 & 58.73 & 72.40 & 61.11 & 62.62 & 58.34 \\
& Cayley-Encoder   & \textbf{61.62} & \textbf{60.38} & \textbf{63.98} & \textbf{64.59} & \textbf{73.94} & \textbf{62.69} & \textbf{64.07} & \textbf{60.32} \\
\midrule
\multirow{2}{*}{Llama3-8B}
& Random-Regular & 54.46 & 44.55 & 48.92 & 46.02 & 65.47 & 55.55 & \textbf{57.68} & 53.51 \\
& Cayley-Encoder   & \textbf{63.05} & \textbf{65.31} & \textbf{63.53} & \textbf{70.17} & \textbf{76.96} & \textbf{63.13} & 55.32 & \textbf{60.74} \\
\bottomrule
\end{tabular}}
\label{tab:ablation_cayley_vs_random_sts}
\end{table*}

\newpage
\section{Cayley Graph Degree}
\label{app:cayley_degree}

To study the effect of degree, we extend $S$ (Appendix~\ref{app:cayley}) while keeping the group fixed. Since a Cayley graph's degree equals $|S|$ and $S$ must be closed under inverses, we add generators in inverse pairs:
$S_6 = S \cup \{AB, (AB)^{-1}\}$ and $S_8 = S_6 \cup \{AB^{-1}, (AB^{-1})^{-1}\}$, where
\[
AB = \begin{pmatrix} 2 & 1 \\ 1 & 1 \end{pmatrix} \qquad
AB^{-1} = \begin{pmatrix} 0 & 1 \\ -1 & 1 \end{pmatrix}
\]
Both have determinant 1 and hence lie in $SL(2,\mathbb{Z}_n)$. For $n \geq 3$ all generators are distinct, so the new resulting graphs are exactly 6 and 8-regular.

\section{Sensitivity to Virtual Nodes}
\label{app:dilution}

When $|V_n| > L$, the Cayley graph contains virtual nodes that are initialized with zero vectors and excluded from the final representation. To test whether a large fraction of virtual nodes dilutes the learned representations, we train Cayley-Encoder with Gemma2-2B (27 layer representations) on Cayley graphs larger than required. Instead of the smallest $n$ such that $|V_n| \geq L$ ($n=4$, 48 nodes, 44\% virtual), we use $n=5$ (120 nodes, 78\% virtual) and $n=6$ (144 nodes, 81\% virtual) with the same hyperparameters. We train on STSBenchmark and evaluate on all STS tasks. Table~\ref{tab:dilution} shows that average performance does not decrease between 44\% and 78\% virtual nodes (63.95 and 64.50) and decreases slightly at 81\% (62.01). This decrease is driven by STS12, STS13, and STS14, while the remaining 5 tasks are unaffected. Since we use the smallest $n$ such that $|V_n| \geq L$, such high virtual-node fractions do not arise for the LLMs we evaluate, except for the smallest Pythia models (71\%), where Cayley-Encoder still outperforms all baselines (Figure~\ref{fig:model_scaling}).

\begin{table}[!htbp]
\centering
\caption{Sensitivity to the fraction of virtual nodes. Cayley-Encoder is trained with Gemma2-2B on Cayley graphs larger than required ($n=4$ is the default) and evaluated on the STS tasks (Spearman correlation $\times 100$). We highlight in bold the best result per column.}
\label{tab:dilution}
\resizebox{\textwidth}{!}{
\begin{tabular}{lccccccccc}
\toprule
Configuration & STSBenchmark & STS12 & STS13 & STS14 & STS15 & STS16 & BIOSSES & SICK-R & Avg \\
\midrule
$n=4$ (48 nodes, 44\% virtual) & 61.62 & 60.38 & 63.98 & \textbf{64.59} & 73.94 & 62.69 & 64.07 & 60.32 & 63.95 \\
$n=5$ (120 nodes, 78\% virtual) & \textbf{61.96} & \textbf{61.08} & \textbf{64.19} & 64.38 & \textbf{74.76} & \textbf{64.13} & \textbf{64.50} & \textbf{61.02} & \textbf{64.50} \\
$n=6$ (144 nodes, 81\% virtual) & 60.56 & 54.84 & 60.36 & 59.28 & 74.03 & 63.51 & 63.38 & 60.14 & 62.01 \\
\bottomrule
\end{tabular}}
\end{table}

\section{DWAtt With and Without Projection}
\label{app:dwatt}

In our main experiments, DWAtt operates on layer representations projected to 256 dimensions. Without the projection, the attention matrices of DWAtt scale with the hidden dimension of the LLM, requiring 69--541$\times$ more trainable parameters than Cayley-Encoder (Table~\ref{tab:dwatt-param-counts}). To evaluate the effect of the projection, we also train DWAtt directly on the full-dimensional layer representations. Tables~\ref{tab:dwatt-sts} and~\ref{tab:dwatt-cls} present the results. The effect of the projection is task-dependent. On classification, the projection improves DWAtt on all four LLMs, and Cayley-Encoder outperforms DWAtt without the projection on all tasks and LLMs. On STS, removing the projection improves DWAtt on all four LLMs, and Cayley-Encoder still achieves a higher average on 3 out of 4 LLMs. On Llama3-8B, DWAtt without the projection achieves a higher average by 1.3 points, while using 484$\times$ more trainable parameters than Cayley-Encoder.

\begin{table}[!htbp]
  \centering
  \footnotesize
  \renewcommand{\arraystretch}{0.95}
  \caption{Trainable parameters of DWAtt with and without the 256-dimensional projection, compared to Cayley-Encoder.}
  \label{tab:dwatt-param-counts}
  \begin{tabular*}{\textwidth}{l @{\extracolsep{\fill}} cccc}
  \toprule
  Method & Pythia-410M & Gemma2-2B & Llama3-8B & Qwen3-8B \\
  \midrule
  Cayley-Encoder & 395K & 722K & 1.18M & 1.18M \\
  DWAtt (projection) & 2.0M & 2.46M & 3.3M & 3.6M \\
  DWAtt (no projection) & 27.4M & 149.0M & 571.1M & 638.3M \\
  \midrule
  \textit{No projection / Cayley-Encoder} & \textit{69$\times$} & \textit{206$\times$} & \textit{484$\times$} & \textit{541$\times$} \\
  \bottomrule
  \end{tabular*}
\end{table}

\begin{table}[!htbp]
\centering
\caption{DWAtt with and without the 256-dimensional projection on STS tasks (Spearman correlation $\times 100$). We highlight in bold the best result per column for each LLM.}
\label{tab:dwatt-sts}
\resizebox{\textwidth}{!}{
\begin{tabular}{llccccccccc}
\toprule
Base Model & Method & STSBenchmark & STS12 & STS13 & STS14 & STS15 & STS16 & BIOSSES & SICK-R & Avg \\
\midrule
\multirow{3}{*}{Pythia-410M} & DWAtt (projection) & 49.4 & 52.5 & 44.6 & 47.5 & 52.5 & 44.8 & 45.3 & 52.4 & 48.6 \\
 & DWAtt (no projection) & 46.2 & \textbf{57.2} & 51.3 & 52.6 & 60.8 & 53.9 & \textbf{65.0} & \textbf{56.9} & 55.5 \\
 & Cayley-Encoder & \textbf{55.8} & 56.0 & \textbf{60.2} & \textbf{56.7} & \textbf{67.0} & \textbf{58.6} & 56.6 & 55.3 & \textbf{58.3} \\
\midrule
\multirow{3}{*}{Gemma2-2B} & DWAtt (projection) & 53.4 & 51.9 & 48.1 & 55.3 & 67.7 & 51.1 & 57.1 & 55.4 & 55.0 \\
 & DWAtt (no projection) & 54.1 & 54.7 & 56.0 & 59.3 & 68.8 & 57.7 & 63.9 & 54.7 & 58.7 \\
 & Cayley-Encoder & \textbf{61.6} & \textbf{60.4} & \textbf{64.0} & \textbf{64.6} & \textbf{73.9} & \textbf{62.7} & \textbf{64.1} & \textbf{60.3} & \textbf{63.9} \\
\midrule
\multirow{3}{*}{Llama3-8B} & DWAtt (projection) & 52.9 & 51.5 & 56.3 & 58.6 & 66.0 & 48.9 & 46.9 & 51.5 & 54.1 \\
 & DWAtt (no projection) & \textbf{66.3} & \textbf{66.4} & 61.8 & \textbf{71.1} & \textbf{79.5} & 62.5 & \textbf{58.9} & \textbf{62.1} & \textbf{66.1} \\
 & Cayley-Encoder & 63.1 & 65.3 & \textbf{63.5} & 70.2 & 77.0 & \textbf{63.1} & 55.3 & 60.7 & 64.8 \\
\midrule
\multirow{3}{*}{Qwen3-8B} & DWAtt (projection) & 38.6 & 45.0 & 41.0 & 40.6 & 54.0 & 45.7 & 54.3 & 42.1 & 45.1 \\
 & DWAtt (no projection) & 36.8 & 58.9 & 31.2 & 34.0 & 55.6 & 49.3 & \textbf{59.5} & 46.8 & 46.5 \\
 & Cayley-Encoder & \textbf{58.7} & \textbf{63.5} & \textbf{64.8} & \textbf{68.9} & \textbf{73.7} & \textbf{58.3} & 54.8 & \textbf{57.9} & \textbf{62.6} \\
\bottomrule
\end{tabular}}
\end{table}

\begin{table}[!htbp]
\centering
\caption{DWAtt with and without the 256-dimensional projection on classification tasks (accuracy). OOM: out of GPU memory; the corresponding average is over the remaining four tasks. We highlight in bold the best result per column for each LLM.}
\label{tab:dwatt-cls}
\resizebox{\textwidth}{!}{
\begin{tabular}{llcccccc}
\toprule
Base Model & Method & Banking77 & Emotion & MTOPDomain & MTOPIntent & PoemSentiment & Avg \\
\midrule
\multirow{3}{*}{Pythia-410M} & DWAtt (projection) & 83.23 & 58.60 & 98.03 & 91.66 & \textbf{70.87} & 80.48 \\
 & DWAtt (no projection) & 84.81 & 57.98 & 98.14 & 88.26 & 61.83 & 78.20 \\
 & Cayley-Encoder & \textbf{89.43} & \textbf{73.83} & \textbf{98.77} & \textbf{94.72} & \textbf{70.87} & \textbf{85.52} \\
\midrule
\multirow{3}{*}{Gemma2-2B} & DWAtt (projection) & 88.82 & 61.22 & 98.39 & 90.93 & 67.79 & 81.43 \\
 & DWAtt (no projection) & 63.20 & 59.11 & 96.06 & 85.42 & 40.10 & 68.78 \\
 & Cayley-Encoder & \textbf{92.58} & \textbf{69.60} & \textbf{99.16} & \textbf{96.43} & \textbf{83.27} & \textbf{88.21} \\
\midrule
\multirow{3}{*}{Llama3-8B} & DWAtt (projection) & 90.04 & 66.55 & 97.97 & 92.41 & 75.00 & 84.39 \\
 & DWAtt (no projection) & OOM & 59.03 & 84.05 & 86.91 & 75.19 & 76.30 \\
 & Cayley-Encoder & \textbf{92.85} & \textbf{73.43} & \textbf{99.03} & \textbf{96.46} & \textbf{79.04} & \textbf{88.16} \\
\midrule
\multirow{3}{*}{Qwen3-8B} & DWAtt (projection) & 91.22 & 68.31 & 98.97 & 92.10 & 73.94 & 84.91 \\
 & DWAtt (no projection) & 85.39 & 59.92 & 97.31 & 73.86 & 63.56 & 76.01 \\
 & Cayley-Encoder & \textbf{92.34} & \textbf{75.70} & \textbf{98.99} & \textbf{95.99} & \textbf{85.38} & \textbf{89.68} \\
\bottomrule
\end{tabular}}
\end{table}

\section{Effectiveness Across Qwen3 Sizes}
\label{app:qwen_sizes}

To complement the Pythia analysis in Section~\ref{sec:scaling} with a second LLM family, we evaluate Last-Layer, FC-Encoder, and Cayley-Encoder on all classification tasks using Qwen3~\citep{yang2025qwen3} with 1.7B, 4B, 8B, and 14B parameters. Table~\ref{tab:qwen-sizes} presents the results. Both encoders outperform Last-Layer by 35--41 percentage points on average at all sizes. Cayley-Encoder outperforms FC-Encoder in 16 out of 20 configurations and on average at every size, with a larger advantage for the two largest LLMs (8B: +2.35, 14B: +1.47) than for the smaller ones (1.7B: +0.74, 4B: +0.31).

\begin{table}[!htbp]
\centering
\caption{Classification accuracy across Qwen3 sizes. Cayley-Encoder outperforms FC-Encoder on average at every size, with a larger margin for the two largest LLMs. We highlight in bold the best result per column for each LLM.}
\label{tab:qwen-sizes}
\resizebox{\textwidth}{!}{
\begin{tabular}{llcccccc}
\toprule
Base Model & Method & Banking77 & Emotion & MTOPDomain & MTOPIntent & PoemSentiment & Avg \\
\midrule
\multirow{3}{*}{Qwen3-1.7B} & Last-Layer & 53.23 & 31.88 & 72.98 & 58.12 & 39.52 & 51.15 \\
 & FC-Encoder & 91.56 & 68.96 & \textbf{98.94} & 94.91 & 79.33 & 86.74 \\
 & Cayley-Encoder & \textbf{92.27} & \textbf{70.01} & 98.87 & \textbf{95.36} & \textbf{80.87} & \textbf{87.48} \\
\midrule
\multirow{3}{*}{Qwen3-4B} & Last-Layer & 55.39 & 31.45 & 74.61 & 62.54 & 39.13 & 52.62 \\
 & FC-Encoder & 91.39 & \textbf{76.05} & 98.60 & 95.51 & 79.62 & 88.23 \\
 & Cayley-Encoder & \textbf{91.78} & 75.91 & \textbf{98.83} & \textbf{95.90} & \textbf{80.29} & \textbf{88.54} \\
\midrule
\multirow{3}{*}{Qwen3-8B} & Last-Layer & 48.12 & 30.18 & 71.79 & 56.67 & 37.31 & 48.81 \\
 & FC-Encoder & 91.90 & 69.66 & \textbf{99.02} & 95.60 & 80.48 & 87.33 \\
 & Cayley-Encoder & \textbf{92.34} & \textbf{75.70} & 98.99 & \textbf{95.99} & \textbf{85.38} & \textbf{89.68} \\
\midrule
\multirow{3}{*}{Qwen3-14B} & Last-Layer & 47.79 & 29.96 & 69.32 & 58.10 & 41.35 & 49.30 \\
 & FC-Encoder & 92.41 & 73.54 & 98.81 & \textbf{96.34} & 80.19 & 88.26 \\
 & Cayley-Encoder & \textbf{92.52} & \textbf{78.68} & \textbf{99.15} & 96.16 & \textbf{82.12} & \textbf{89.73} \\
\bottomrule
\end{tabular}}
\end{table}

\section{Hardware Details}
\label{app:hardware}

All experiments were conducted on an HPC cluster with access to NVIDIA A6000, A100, V100, and GeForce RTX 3090 GPUs. We do not report training time comparisons, as experiments were distributed across different GPU types and scheduling queues, making direct timing comparisons unreliable.

\section{Asset Licenses and Terms of Use}
\label{app:licenses}

All models and datasets used in this work are publicly available and used in accordance with their respective licenses.

\subsection{Model Licenses}
The following large language models served as the frozen backbones for our experiments \citep{biderman2023pythia,gemma_2024, llama3modelcard, yang2025qwen3}  :
\begin{itemize}
    \item \textbf{Pythia Suite (14M--2.8B)}: Licensed under the \textbf{Apache License 2.0}.
    \item \textbf{Gemma2-2B}: Released under the \textbf{Gemma Community License}.
    \item \textbf{Llama3-8B}: Released under the \textbf{Meta Llama 3 Community License Agreement}.
    \item \textbf{Qwen3 (1.7B--14B)}: Licensed under the \textbf{Apache License 2.0}.
\end{itemize}

\subsection{Dataset Licenses (MTEB Benchmark)}
The datasets used for evaluation were accessed via the Massive Text Embedding Benchmark (MTEB) \citep{muennighoff2023mteb} which is licensed under the \textbf{Apache License 2.0}. Individual dataset licenses are summarized below:

\begin{table}[h]
\centering
\begin{tabular}{lll}
\hline
\textbf{Task Category} & \textbf{Dataset} & \textbf{License} \\ \hline
\textbf{Classification} & Banking77 & \textbf{CC BY 4.0} \\
 & Emotion & \textbf{CC BY 4.0} \\
 & MTOP (Domain/Intent) & \textbf{CC BY-SA 4.0} \\
 & Poem Sentiment & \textbf{Apache License 2.0} \\ \hline
\textbf{STS} & STSBenchmark & \textbf{CC BY-SA 4.0} \\
 & STS12--16 & \textbf{CC BY-SA 4.0} \\
 & BIOSSES & \textbf{CC BY-NC 4.0} \\ 
 & SICK-R & \textbf{CC BY-NC-SA 4.0} \\ \hline 
\end{tabular}
\caption{Verified licenses for the datasets used in this work.}
\label{tab:dataset_licenses}
\end{table}

\newpage

\end{document}